\documentclass[10pt,twocolumn,letterpaper]{article}

\usepackage{cvpr}              % To produce the CAMERA-READY version
\definecolor{cvprblue}{rgb}{0.21,0.49,0.74}
\usepackage[pagebackref,breaklinks,colorlinks,allcolors=cvprblue]{hyperref}

\usepackage[utf8]{inputenc} % allow utf-8 input
\usepackage[T1]{fontenc}    % use 8-bit T1 fonts
\usepackage{hyperref}       % hyperlinks
\usepackage{url}            % simple URL typesetting
\usepackage{booktabs}       % professional-quality tables
\usepackage{amsfonts}       % blackboard math symbols
\usepackage{nicefrac}       % compact symbols for 1/2, etc.
\usepackage{microtype}      % microtypography
\usepackage{xcolor}         % colors
\usepackage{graphicx} 
\usepackage{wrapfig}
\usepackage{amsmath}
\usepackage{subcaption}
\usepackage{caption}
\usepackage{array}           % For \arraystretch
\usepackage{threeparttable}  % For threeparttable environment
\usepackage[table]{xcolor}   % For \rowcolor
\usepackage{makecell}        % For multi-line cells (alternative to \tabincell)
\usepackage{pifont}
\usepackage{multirow}
\usepackage{tabularx}
\usepackage{etoolbox}
\usepackage{algorithm}
\usepackage{algpseudocode}
\usepackage[most]{tcolorbox}
\usepackage{lipsum} 
\usepackage{fontawesome5}

\newtheorem{lemma}{Lemma}
\newtheorem{assumption}{Assumption}
\newtheorem{proof}{Proof}[section]

\newtcolorbox{theorembox}{
  colback=blue!20!cyan!10!white!50,
  colframe=gray!100!black,
  arc=1mm, 
  boxrule=1pt,  
  left=5pt,right=5pt,top=5pt,bottom=5pt, 
  width=0.99\linewidth 
}
\newenvironment{takeaway}[1][]{
  \begin{theorembox}
    \itshape
}{
  \end{theorembox}
}
\def\paperID{*****} % *** Enter the Paper ID here
\def\confName{CVPR}
\def\confYear{2026}

\title{Test-Time Distillation for Continual Model Adaptation}

\author{
Xiao Chen$^1$\footnotemark[1],  
Jiazhen Huang$^{1}$\footnotemark[1],
Zhiming Liu$^1$,  \\
Qinting Jiang$^1$, 
Fanding Huang$^1$, 
Jingyan Jiang$^2$\footnotemark[2], 
Zhi Wang$^1$\footnotemark[2]
\\
$^1$ Shenzhen International Graduate School, Tsinghua University
$^2$ Shenzhen Technology University \\
}

\begin{document}
\maketitle

\renewcommand{\thefootnote}{\fnsymbol{footnote}}
\footnotetext[1]{Equal contribution.}
\footnotetext[2]{Co-corresponding authors.  \faEnvelope~ \small{ \texttt{jiangjingyan@sztu.edu.cn; wangzhi@sz.tsinghua.edu.cn}}.}

\renewcommand{\thefootnote}{\arabic{footnote}}

\begin{abstract}
Deep neural networks often suffer performance degradation upon deployment due to distribution shifts. 
Continual Test-Time Adaptation (CTTA) aims to address this issue in an unsupervised manner. 
However, existing methods that rely on self-supervision are prone to an inherent self-referential feedback loop that amplifies initial prediction errors, leading to model drift. 
We revisit this limitation and propose Test-Time Distillation (TTD), which reframes adaptation as a distillation process guided by a frozen Vision-Language Model (VLM) as an external signal. 
While promising, we find that direct distillation is fraught with two pitfalls: (1) the Generalist Trap, where the VLM's broad but non-specialized knowledge leads to suboptimal performance on specific tasks and shifts; and (2) the Entropy Bias, where naive model fusion techniques based on entropy fail due to the disparate calibration of heterogeneous models. 
These pitfalls highlight the need to build a robust supervisory signal and leverage it to guide the target model toward stable adaptation. 
Hence, we present \textbf{CoDiRe}, a \textbf{Co}ntinual \textbf{Di}stillation and \textbf{Re}ctification framework for TTD. 
CoDiRe first constructs a robust blended teacher by dynamically fusing the predictions of the VLM and the target model. 
Critically, it circumvents the Entropy Bias by leveraging Maximum Softmax Probability (MSP) as a more reliable confidence metric for weighting each model's expertise. 
Then applies an Optimal Transport-based rectification to further align predictions with the blended teacher, enabling continuous and stable adaptation. 
Extensive experiments show that CoDiRe outperforms state-of-the-art baselines, exceeding CoTTA by 10.55\% with only 48\% of its time cost on ImageNet-C. 
Project page is publicly available at \url{https://github.com/walawalagoose/TTD}.
\end{abstract}
\vspace{-20pt}

\section{Introduction}
\label{sec:intro}

%%%%%%%%%%%%%%%%%%%%%%%%%Introduction%%%%%%%%%%%%%%%%%%%%%%%
\begin{figure}
    \vspace{-15pt}
    \centering
    \includegraphics[width=0.9\linewidth]{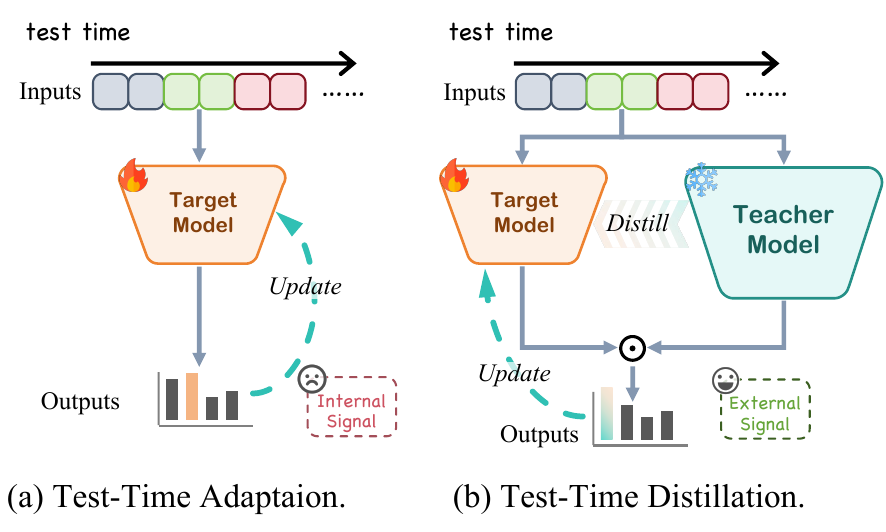}
    % \captionsetup{width=0.99\linewidth}  % Set caption width to match image width
    \vspace{-10pt}
    \caption{
    \textbf{Comparison between TTA and TTD.}
    (a) TTA updates the source-pretrained target model solely based on the internal signals with a self-supervised loss.
    (b) TTD introduces a VLM as a teacher model to provide external signals during inference.
}
    \label{fig:intro}
    \vspace{-20pt}
\end{figure}
% \par  % End the paragraph and reset wrapping
% \noindent  % Prevent indentation of the next line

%%%%%%%%%%%%%%% INTRO TO TTA & CTTA %%%%%%%%%%%%%%%%%%
Deep neural networks (DNNs) \cite{DNN} frequently encounter deployment environments that deviate from their training distributions, leading to degraded performance and reliability. 
Test-Time Adaptation (TTA) \cite{Tent,bn_adapt,TTAB, jiang2025feature} therefore aims to align models to the target distribution on the fly, without access to source data or offline fine-tuning. 
Among these methods, Continual Test-Time Adaptation (CTTA)~\cite{cotta} enables sequential adaptation across evolving distribution shifts.
The prevailing CTTA approach, from CoTTA \cite{cotta} to more recent continual variants \cite{santa, rotta, vida}, is rooted in self-supervision. 
These methods rely exclusively on the model's own predictions as an \textit{internal signal} to generate learning targets, typically through self-distillation.

However, we argue that this reliance on a self-referential signal constitutes a limitation. 
Under a significant domain shift, the model's initial predictions are inherently noisy and unreliable. 
Employing such outputs as supervisory signals creates a risky feedback loop: initial errors might be amplified rather than corrected, leading to gradual drift away from the optimum. 
The core problem of CTTA is therefore not merely adapting, but adapting reliably without reinforcing its own biases. 
This motivates our central question: \textit{Can we leverage external knowledge to construct a stable and robust anchor signal that breaks this error accumulation cycle and guides the model towards a reliable optimization direction?}

To identify such an external anchor, we leverage the vast open-world knowledge encapsulated by modern Vision-Language Models (VLMs) \cite{vlm2, openclip, siglip}, with CLIP~\cite{clip} as a representative exemplar. 
Pre-trained on web-scale image-text pairs, CLIP develops a rich semantic understanding that is orthogonal to the inductive biases of any single, task-specific classifier. 
Therefore, CLIP is independent of the source training set and, crucially, is not influenced by the target domain shift affecting the target model. 
This makes CLIP an ideal candidate to serve as a signal source for adaptation. 
Therefore, we introduce a new paradigm, \textbf{Test-Time Distillation (TTD)}, which reframes adaptation as a distillation process guided by a frozen VLM teacher,  shifting from self-referential correction to external guidance.

\begin{figure}
    \vspace{-15pt} 
    \centering
    \includegraphics[width=0.85\linewidth]{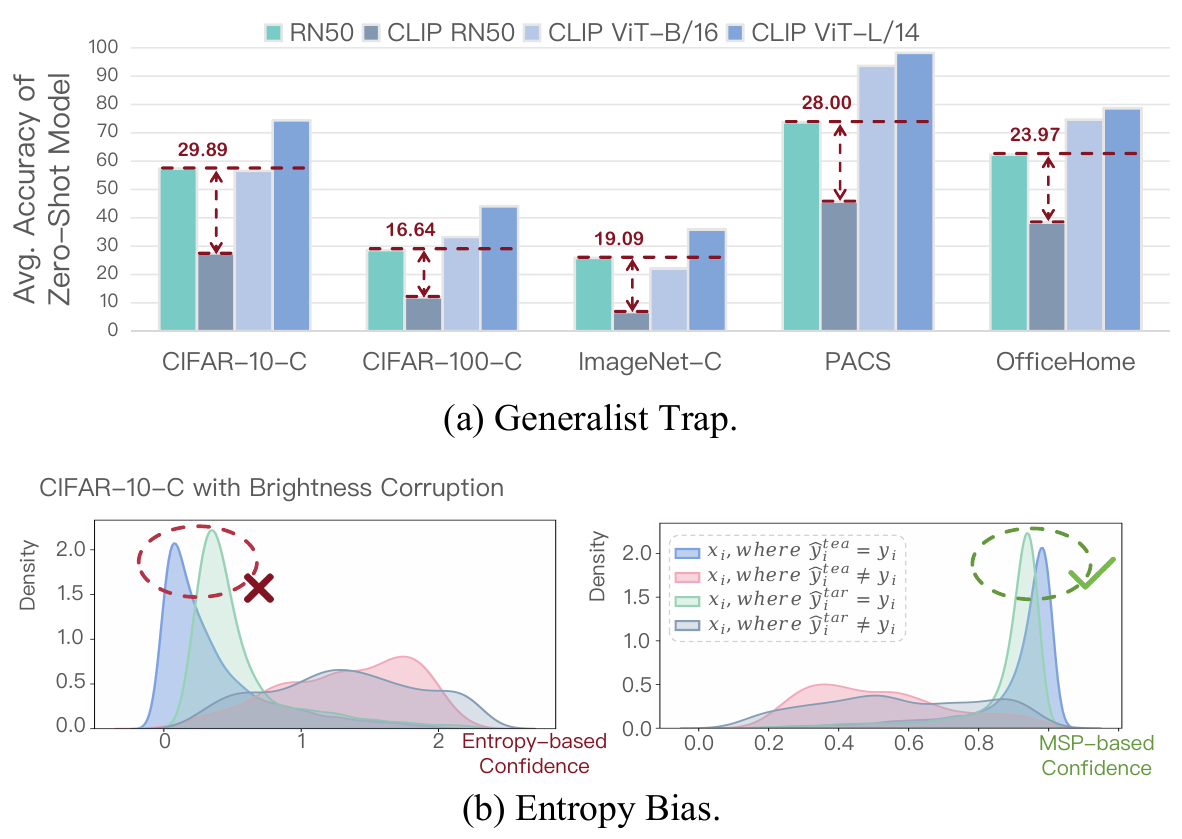}
    \captionsetup{width=\linewidth}  % Set caption width to match image width
    \vspace{-10pt}
    \caption{
    \textbf{Two pitfalls of Test-Time Distillation.}
    (a) \textit{Generalist Trap.} Under domain shift, CLIP underperforms a supervised classifier with the same backbone.
    (b) \textit{Entropy Bias.} Heterogeneous models exhibit inconsistent entropy distributions, inducing bias in entropy-based confidence.
}
    \label{fig:emperical_pitfalls}
    \vspace{-0.5cm}  
\end{figure}

Although CLIP provides an appealing source of external supervision for stabilizing TTA, the central challenge lies in transforming its broad, open-world knowledge into a dependable and task-relevant signal for the target model. 
Although direct distillation from CLIP appears promising in principle, it often suffers from two critical pitfalls in practice: (1) \textit{Generalist Trap}: 
While CLIP possesses broad semantic knowledge, its generalist nature makes it vulnerable under domain shifts and thus not always effective.
As shown in Figure~\ref{fig:emperical_pitfalls}(a), CLIP consistently underperforms supervised classifiers with the same backbone on task-specific benchmarks, and struggles when facing certain shifts like common corruptions. 
Notably, scaling does not help much to bridge this gap. On ImageNet-C, even CLIP ViT-L/14 (304M parameters, more than $10\times$ larger than RN50 with 23M) only achieves about 10\% performance gain over RN50 source-pretrained on ImageNet, and still trails the source-pretrained ViT-B/16 (as illustrated in Section \ref{sec:exp-main}).
Recent studies~\cite{zhang2024out, zhang2024visually} have further revealed that scaling CLIP size is insufficient for reliable out-of-distribution (OOD) generalization. 
This suggests that neither CLIP alone nor the target model alone offers a sufficiently reliable supervisory signal. 
To escape this pitfall, we aim to apply certain model fusion techniques to integrate the knowledge from both models to form a more robust signal.
(2) \textit{Entropy Bias}: Determining a reliable fusion technique introduces a new challenge. 
Recent works have adopted entropy-based confidence~\cite{oh2024dawin,khan2025slidemambaentropybasedadaptivefusion,simons2023summitsourcefreeadaptationunimodal,chen2023easynet} as a proxy for model expertise and thus as weights for model merging, feature fusion, or ensembling.
However, as illustrated in Figure~\ref{fig:emperical_pitfalls}(b), we find that heterogeneous models exhibit distinct entropy landscapes due to architectural, calibration, and training differences, making entropy inherently biased toward models with globally lower entropy and leading to skewed distillation targets. 
Hence, our core objective has shifted to: \textit{how to build a robust supervisory signal from an effective fusion technique, and leverage it as teacher to guide the target model toward stable adaptation?}

%%%%%%%%%%%%%%%%%%%%%%%%%% METHOD DESCRIPTION %%%%%%%%%%%%%%%%%%%%%%%%%%%%%
 To this end, we propose \textbf{CoDiRe}, a \textbf{Co}ntinual \textbf{Di}stillation and \textbf{Re}ctification framework for CTTA, which combines a frozen CLIP and the target model to construct a robust teacher signal for TTD. 
 CoDiRe comprises two components: Distillation and Rectification. 
 In Distillation, we interpolate the logits of CLIP and the target model to ensemble a robust blended teacher  as a robust supervisory signal. 
 It sidesteps Entropy Bias by utilizing confidence score based on Maximum Softmax Probability (MSP), which proves to be a more robust arbiter between heterogeneous models.
 This allows it to intelligently fuse the target model and VLM predictions into a high-quality signal. 
In Rectification, we further refine the target model by deriving a rectification matrix using optimal transport (OT)~\cite{ot1, ot2} to align predictive mass with the target label geometry under constraints imposed by the blended teacher.
We validate CoDiRe across diverse CTTA benchmarks and shift scenarios, demonstrating that it not only surpasses various TTA baselines based on target model, but also outperforms several variants based on CLIP. 
Our contributions are summarized as follows:
\begin{itemize}
\item \textbf{New Paradigm}: We propose Test-Time Distillation, \textit{first} introducing a VLM as a distillation teacher into CTTA.
\item \textbf{Empirical Findings}: We identify two practical pitfalls under TTD paradigm: Generalist Trap and Entropy Bias.
\item \textbf{Novel Methodology}: We develop a TTD method CoDiRe, which constructs a more reliable blended teacher and mitigates the impacts of the aforementioned pitfalls.
\item \textbf{Extensive Experiments}: We demonstrate substantial and robust gains of our CoDiRe over state-of-the-art baselines across CTTA benchmarks, surpassing CoTTA by 10.55\% while requiring only 48\% of its time cost on ImageNet-C.
\end{itemize}
\par  % End the paragraph and reset wrapping
\noindent  % Prevent indentation of the next line

\section{Related Works}
\paragraph{Test-Time Adaptation.}
TTA mitigates performance degradation on OOD data by fine-tuning a pretrained model during inference using mini-batches in an online manner~\cite{shift_1,shift_2}. 
Tent~\cite{Tent} first introduced the concept of Fully Test-Time Adaptation, proposing entropy-based self-supervised optimization, which was subsequently adopted by follow-up work~\cite{Memo, SAR, EATA, DEYO, cliff, chen2025neural, fan2025moetta}. 
As the field has evolved, TTA has been extended to diverse modalities and tasks, including depth completion~\cite{park2024test}, action recognition~\cite{xiong2024modality}, point cloud understanding~\cite{jiang2024pcotta}, and time-series anomaly detection~\cite{kim2024model}. 
Recent works~\cite{tpt, tda, boostadapter, huang2025cosmic} further extend TTA to VLMs for improved OOD generalization. 
Building on this foundation, CoTTA~\cite{cotta} introduced CTTA to handle continuously shifting test distributions, mitigating catastrophic forgetting via self-distillation with the source model as the teacher. 
Subsequent methods~\cite{rotta, santa} further advanced this line of work; however, whether employing entropy-based self-supervised optimization or self-distillation using the source model as the teacher, the supervisory signal is inherently internal, thereby imposing an intrinsic performance ceiling. 
Instead, we propose the TTD paradigm, which employs CLIP as a teacher to form a reliable and robust external signal.

\paragraph{Vision-Language Models (VLMs).}
VLMs learn joint visual-textual representations that enable open-vocabulary understanding and zero-shot transfer across diverse tasks. 
Early methods, such as bottom-up top-down attention~\cite{anderson2018bottom}, BAN~\cite{kim2018bilinear}, and MCAN~\cite{yu2019deep}, made notable progress on vision-language benchmarks, while recent models including BLIP~\cite{vlm4}, MiniGPT4~\cite{minigpt4}, and BERT-based architectures~\cite{lu2019vilbert} further advanced multimodal reasoning. 
Among them, CLIP~\cite{clip} popularized contrastive pretraining and has shown strong performance across 3D~\cite{zhang2023learning}, video~\cite{lin2022frozen}, and depth understanding~\cite{zhang2022can}. 
In this work, we adopt CLIP as the teacher model, given its broad recognition and widespread use. 
However, CLIP can underperform on OOD data~\cite{tpt,tda}, and its performance can lag behind that of a supervised classifier with the same backbone--a phenomenon we term the Generalist Trap. 
This motivates our attempt at constructing a more reliable teacher.

\paragraph{Knowledge Distillation.}
Knowledge distillation trains lightweight student models under the supervision of large pre-trained teachers, achieving notable success across tasks such as visual recognition~\cite{huang2024etag, li2023curriculum, yang2022mixskd} and multimodal representation learning~\cite{fang2021compressing, li2024correlation, zhang2022unims}. To enable effective transfer, diverse paradigms have emerged, including feature imitation~\cite{wu2023tinyclip, yang2024clip}, relational distillation~\cite{yang2022mutual, yang2023online}, and prompt distillation~\cite{li2024promptkd, wu2024cascade}. CLIPPING~\cite{pei2023clipping} introduces a hierarchical alignment strategy that promotes student-centric adaptation for efficient assimilation of teacher knowledge while CLIP-KD~\cite{yang2024clip} transfers knowledge by minimizing unimodal feature discrepancies between student and teacher. However, these works primarily focus on training a compact network via offline distillation. In contrast, we propose TTD, wherein the target model self-evolves within the test stream under guidance from CLIP, integrating rich world knowledge while preserving its own distributional sensitivity.

\section{Pitfalls in Test-Time Distillation}
\label{sec:empirical}

\begin{figure}
    \vspace{-15pt}
    \centering
    \includegraphics[width=0.75\linewidth]{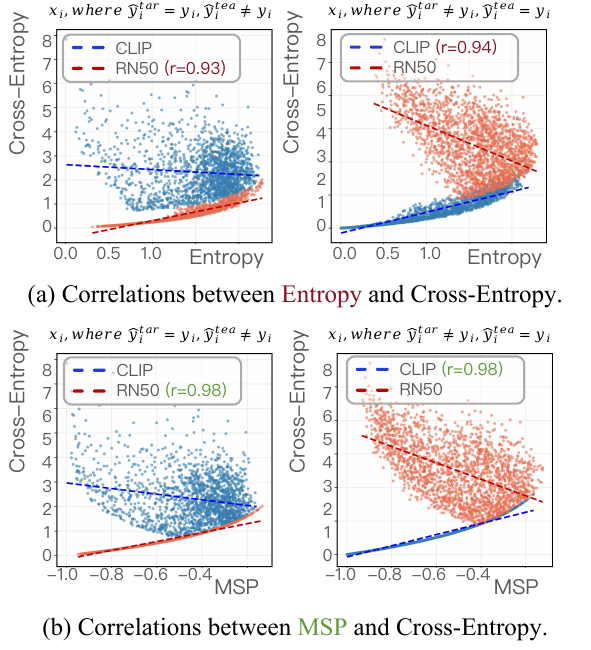}
    \label{fig:scatter_conf_bottom}
    \captionsetup{width=\linewidth}  % Set caption width to match image width
    \vspace{-10pt}
    \caption{
    \textbf{Correlations between different proxies and cross-entropy.}
    MSP-based confidence shows a substantially stronger correlation than entropy-based confidence, indicating that MSP is a more reliable proxy for cross-entropy.
}
    \label{fig:empirical_correlations}
    \vspace{-18pt}
\end{figure}
\par  % End the paragraph and reset wrapping
\noindent  % Prevent indentation of the next line
We introduce TTD as a new paradigm, with CLIP as an external signal to construct the distillation teacher objective. 
Nevertheless, we identify two pitfalls in practice: the Generalist Trap precludes directly using CLIP as the distillation target, motivating the construction of a more reliable blended teacher; and the Entropy Bias invalidates entropy-based confidence as a cross-model expertise measure for fusion, thereby prompting the exploration for alternative solutions as weights to fuse the two models.

\paragraph{Pitfall 1: Generalist Trap.}
TTD is proposed to address the limitations of self-distillation in existing CTTA methods~\cite{cotta, santa, rotta}, where reliance on the internal signals to generate learning targets constrains performance. 
A naive method is to directly adopt the output of CLIP as the teacher signal, but empirical evidence indicates this does not consistently improve the target model due to the Generalist Trap:
\begin{takeaway}
\textbf{Generalist Trap.} Under domain shift, CLIP underperforms a supervised classifier with the same backbone.
\end{takeaway}
As shown in Figure~\ref{fig:emperical_pitfalls}(a), on canonical domain generalization (PACS and OfficeHome) and synthetic corruption (CIFAR-10-C and ImageNet-C) benchmarks, CLIP RN50 consistently underperforms an RN50 classifier trained on the source domain. 
Moreover, CLIP particularly struggles on certain shifts like common corruptions (e.g., noise, blur). 
Even with a stronger backbone, CLIP ViT-B/16 remains inferior to RN50 on the two corruption datasets. 

To remedy this pitfall, two avenues exist: upgrading the teacher model or constructing a more reliable distillation target. 
However, existing studies~\cite{zhang2024out, zhang2024visually} demonstrate that: 1) Proprietary and public VLMs such as LLaVA~\cite{liu2023llava}, BLIP-2~\cite{vlm5}, and GPT-4-Turbo~\cite{achiam2023gpt}, despite often employing CLIP~\cite{clip} as the vision encoder and possessing substantially more parameters, significantly underperform CLIP on standard image classification tasks; 2) Scaling up CLIP on OOD datasets does not bring remarkable benefit and sometimes can even hurt. 
Therefore, we pursue the latter path of constructing a more reliable distillation target called blended teacher.
Inspired by Ensemble Learning \cite{ensemble1,ensemble2}, we aim to simply fuse the output logits of CLIP and the target model to mitigate performance disparities.

\begin{figure*}[!h]
    \centering
    \vspace{-20pt}
    \includegraphics[width=0.8\textwidth]{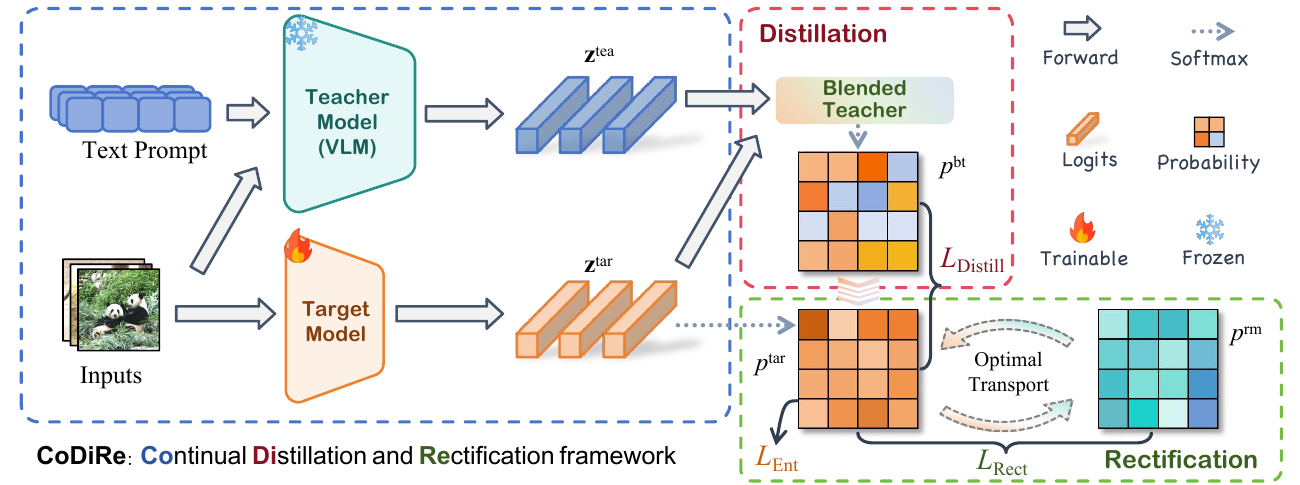}
    \vspace{-8pt}
    \caption{\textbf{Overview of our proposed CoDiRe}.
    CoDiRe introduces a VLM to provide external signals and consists of two components-Distillation and Rectification-by constructing a reliable blended teacher to mitigate the effects of Generalist Trap and Entropy Bias.
  }
  \vspace{-10pt}
    \label{fig:pipeline}
\end{figure*}
\paragraph{Pitfall 2: Entropy Bias.}
The blended teacher is expected to provide a stronger supervisory signal by integrating the knowledge from the two models. 
Beyond merely averaging, an intuitive idea is to assign higher weights to the model that performs better on the current data.
This requires a metric to measure each model's expertise. 
The cross-entropy (CE) between predictions and the ground truth naturally becomes an oracle choice, since a lower CE indicates higher prediction accuracy. 
However, it is impossible in the unsupervised TTA setting as labels are unavailable. 
A common workaround is to utilize some kinds of confidence measure as proxies for CE, as is widely done with entropy in ensembling~\cite{simons2023summitsourcefreeadaptationunimodal}, feature fusion~\cite{zou2025zeroshotsubjectcentricgenerationcreative, chen2023easynet} and model merging~\cite{oh2024dawin}. 
However, our experiments reveal an Entropy Bias in TTD:
\begin{takeaway}
\textbf{Entropy Bias.} Heterogeneous models exhibit inconsistent entropy distributions, inducing bias in entropy-based confidence.
\end{takeaway}
As illustrated in Figure~\ref{fig:emperical_pitfalls}(b), the peak of CLIP's entropy distribution shows a clear deviation from that of the target model, causing entropy-based confidence to skew the fusion results toward models with globally lower entropy. 
Tuning the temperature parameter can adjust CLIP’s logit sharpness, yet we avoid learning or hand-crafting a task-specific temperature for every scenario. 
Instead, we seek a more reliable metric in this setting with two properties: 1) effectiveness (strong correlation with ground-truth model expertise); and 2) comparability (distributional consistency across heterogeneous models without pronounced shifts). 

Through experiments, we identify Maximum Softmax Probability (MSP) as a superior proxy for CE. 
In Figure~\ref{fig:emperical_pitfalls}(b), MSP-based confidence substantially mitigates inter-model confidence bias, providing a more faithful confidence metric than entropy. 
Furthermore, in Figure~\ref{fig:empirical_correlations}, we demonstrate the correlation plots between (MSP, CE) and (entropy, CE), focusing on cases where the target model and CLIP predictions disagree. 
The results show that under prediction conflicts, MSP-based confidence maintains a significantly higher correlation with cross-entropy than entropy. 
Consequently, an MSP-driven blended teacher better balances model predictions, yielding a more robust distillation target. 
We further provide a theoretical analysis motivated by an empirical MSP-accuracy binning experiment in the Supplementary Materials to show the effectiveness of this choice.

\section{Methodology}
\label{sec:method}
The overview of our proposed CoDiRe is depicted in Figure~\ref{fig:pipeline}. 
CoDiRe comprises two core components: Distillation (Section~\ref{sec:method-distill}), which constructs a more reliable blended teacher for distillation; and Rectification (Section~\ref{sec:method-rect}), which further refines the target model under constraints imposed by the blended teacher. 
In addition, we incorporate a widely adopted entropy loss and conduct a systematic study of reset mechanisms in the CTTA setting (Section~\ref{sec:method-overall}). 

\subsection{Preliminaries}
Given a pretrained classifier \(f(\cdot)\) with parameters \(\theta_0\) trained on source data \((\mathcal{X}^S, \mathcal{Y}^S)\), we seek to enhance the performance of this target model during inference for a continually evolving target domain in an online manner, aided by a teacher model \(\mathcal{F}(\cdot)\), such as CLIP. 
Unlabeled data \(\mathcal{X}^T\), comprising \(K\) classes from the target domain, arrive sequentially and asynchronously, and the model has access only to the data available at the current time step. 
At time step \(t\), the model receives a mini-batch of unlabeled test samples \(\{x \mid x \in \mathcal{X}^T_t\}\), and the target model \(f(\cdot)\) must adapt its parameters for future inputs, i.e., \(\theta_t \rightarrow \theta_{t+1}\). 
The distribution \(\mathcal{X}^T_t\) evolves continually over time. 
For a test sample \(x_i\), the model produces logits \(\mathbf{z}_i\), with post-softmax probabilities \(p_i=\sigma(\mathbf{z}_i)\), where \(\sigma(\cdot)\) denotes the softmax operator. Specifically, for \(x_i\), we denote \(\mathbf{z}^{\text{tar}}_i=f(x_i)\) and \(\mathbf{z}^{\text{tea}}_i=\mathcal{F}(x_i)\). 
We form the blended teacher via a linear interpolation with weight \(\lambda_i\):
\begin{equation}
    \mathbf{z}^{\text{bt}}_i = \lambda_i \cdot \mathbf{z}^{\text{tea}}_i + (1-\lambda_i) \cdot \mathbf{z}^{\text{tar}}_i,
\end{equation}
and adopt \(p^{\text{bt}}_i\) as the inference prediction output, where \(p^{\text{bt}}_i = \sigma(\mathbf{z}^{\text{bt}}_i)\).
To ensure balance and scale consistency, both logits are independently normalized by subtracting its LogSumExp (LSE) term and thus operate in the log-probability space.

\subsection{Distillation}
\label{sec:method-distill}
Section~\ref{sec:empirical} shows that directly distilling from CLIP is suboptimal, and entropy-based interpolation of the target model and CLIP predictions is likewise ineffective. 
Empirical evidence indicates that MSP is a superior proxy for CE. 
To this end, we construct the blended teacher via MSP-based confidence and use the resulting distribution as the distillation target. The MSP-based confidence weight \(\lambda_i\) is computed as:
\begin{equation}
\lambda_i = \frac{\exp(\max(p^{\text{tea}}_i))}{\exp(\max(p^{\text{tea}}_i))+\exp(\max(p^{\text{tar}}_i))}.
\label{eq:fusion}
\end{equation}
We also explore other weighting schemes such as naive averaging and entropy-based alternatives, detailed in Section~\ref{sec:discussion}.
The MSP-based blended teacher serves as a superior distillation target compared to CLIP, effectively mitigating the impacts of the pitfalls of the Generalist Trap and Entropy Bias in TTD. 
Intuitively, higher-confidence blended teachers should also exert greater influence. We therefore weight the distillation loss by the teacher’s confidence. 
The distillation loss \(\mathcal{L}_{\text{Distill}}\) is then computed as:
\begin{equation}
\mathcal{L}_{\text{Distill}}(x_i) = -\,\max(p^{\text{bt}}_i)\sum_{c=1}^{K} p_{ic}^{\text{bt}} \log p_{ic}^{\text{tar}}.
\end{equation}

\begin{table*}[ht]
\caption{\textbf{Comparison results under corruption scenarios on CIFAR-10-C.} Classification accuracy of the standard CIFAR-10 \(\rightarrow\) CIFAR-10-C online continual test-time adaptation task while continually adapting to different corruptions at the highest severity 5. The best and second-best results are highlighted in \textbf{bold} and \underline{underlined}, respectively.}
\vspace{-10pt}
\centering
\resizebox{0.95\textwidth}{!} {
\begin{tabular}{c|c|ccccccccccccccc|c}
\toprule
\textbf{CIFAR-10-C} &
\multirow{2}{*}{\textbf{Venue}} &
\multicolumn{3}{c}{\textbf{Noise}} &
\multicolumn{4}{c}{\textbf{Blur}} &
\multicolumn{4}{c}{\textbf{Weather}} &
\multicolumn{4}{c|}{\textbf{Digital}} &
\multirow{2}{*}{\textbf{Avg.}} \\ %\cline{2-16}
\textbf{RN50} & & \textit{gauss.} & \textit{shot} & \textit{impul.} & \textit{defoc.} & \textit{glass} & \textit{motion} & \textit{zoom} & \textit{snow} & \textit{frost} & \textit{fog} & \textit{brit.} & \textit{contr.} & \textit{elastic} & \textit{pixel} & \textit{jpeg} &  \\ 
\rowcolor{gray!10} & & \multicolumn{15}{c|}{\textit{\small{Time}}\hspace{2em}\textemdash\textemdash\textemdash\textemdash\textemdash\textemdash\textemdash\textemdash\textemdash\textemdash\textemdash\textemdash\textemdash\textemdash\textemdash\textemdash\textemdash\textemdash\textemdash\textemdash\textemdash\textemdash\textemdash\textemdash\textemdash\textemdash\textemdash\textemdash\textemdash\textemdash\textemdash\textemdash\textemdash\textemdash\textemdash\textemdash\textemdash\textemdash\textemdash\textemdash\textemdash\textemdash>} & \\ 
\midrule
\textbf{Source}
  & - & 33.69 & 40.07 & 35.74 & 53.11 & 49.86 & 67.04 & 56.85
  & 73.89 & 62.59 & 64.66 & 89.00 & 43.57 & 74.37 & 41.96 & 74.43 & 57.39 \\

\textbf{BN Adapt}
  & NIPS'20 & 62.43 & 64.33 & 57.30 & 84.48 & 59.46 & 82.02 & 83.25
  & 76.87 & 74.30 & 79.29 & 87.24 & 83.33 & 71.72 & 73.41 & 71.10 & 74.03\(_{\pm0.05}\) \\

\textbf{Tent}
  & ICLR'21 & 64.70 & 70.66 & 66.39 & \underline{86.66} & 66.20 & \underline{85.09} & 86.37
  & 82.10 & 81.57 & \underline{83.29} & 90.15 & 85.03 & 78.90 & 81.40 & 79.19 & 79.18\(_{\pm0.11}\) \\

\textbf{MEMO}
  & NIPS'22 & 50.07 & 56.89 & 52.72 & 70.75 & 58.65 & 76.79 & 71.30
  & 79.74 & 74.47 & 74.08 & \underline{90.72} & 65.89 & 77.26 & 49.76 & 77.27 & 68.42\(_{\pm0.03}\) \\

\textbf{EATA}
  & ICML'22 & 62.36 & 64.34 & 57.31 & 84.49 & 59.46 & 82.02 & 83.25
  & 76.87 & 74.31 & 79.29 & 87.24 & 83.33 & 71.71 & 73.43 & 71.11 & 74.03\(_{\pm0.05}\) \\

\textbf{SAR}
  & ICLR'23 & 63.73 & 68.82 & 63.33 & 85.65 & 64.34 & 83.20 & 85.35
  & 79.91 & 79.09 & 81.42 & 88.75 & 83.93 & 77.00 & 79.40 & 77.75 & 77.44\(_{\pm0.07}\) \\

\textbf{DeYO}
  & ICLR'24 & \underline{67.70} & \underline{76.27} & 70.98 & 86.48 & \underline{70.04} & 84.59 & \underline{87.10}
  & \underline{82.25} & 82.44 & 82.72 & 89.49 & \underline{86.28} & \underline{79.19} & \underline{82.69} & \underline{80.78} & \underline{80.60}\(_{\pm0.07}\) \\
\midrule

\textbf{CoTTA}
  & CVPR'22 & 61.57 & 63.26 & 61.05 & 84.52 & 53.21 & 81.93 & 83.28
  & 76.82 & 74.23 & 79.25 & 87.21 & 83.30 & 71.59 & 73.32 & 70.97 & 73.70\(_{\pm0.03}\) \\

\textbf{NOTE}
  & NIPS'22 & 55.11 & 68.63 & 55.80 & 34.33 & 56.29 & 69.34 & 78.37 & 72.35 & 78.81 & 64.58 & 86.97 & 73.04 & 63.46 & 50.32 & 70.17 & 65.17\(_{\pm0.07}\) \\
  
\textbf{RoTTA}
  & CVPR'23 & 54.09 & 53.76 & 50.24 & 76.77 & 53.15 & 76.76 & 80.39 & 71.82 & 61.30 & 73.48 & 84.82 & 43.55 & 64.42 & 72.48 & 73.94 & 66.06\(_{\pm0.50}\) \\
  
\textbf{SANTA}
 & TMLR'23 & 65.21 & 70.75 & 65.18 & 85.21 & 65.67 & 82.88 & 84.44
  & 80.31 & 79.41 & 81.36 & 89.13 & 83.75 & 77.08 & 80.27 & 78.60 & 77.95\(_{\pm0.28}\) \\  
  
\textbf{ViDA}
 & ICLR'24 & 62.53 & 64.71 & 57.95 & 84.67 & 60.57 & 82.42 & 83.85
  & 77.79 & 75.77 & 80.17 & 87.89 & 83.50 & 73.56 & 75.39 & 73.38 & 74.94\(_{\pm0.08}\) \\

\midrule

\textbf{CLIP}
  & ICML'21 & 65.39 & 66.64 & \underline{76.02} & 75.75 & 48.18 & 78.16 & 79.08
  & 81.92 & \underline{84.68} & 76.14 & 90.40 & 80.47 & 64.74 & 76.93 & 71.32 & 74.39 \\

\rowcolor{cyan!10}
\textbf{Ours}
 & - & \textbf{79.26} & \textbf{84.45} & \textbf{84.36} & \textbf{89.68} & \textbf{74.79} & \textbf{89.50} & \textbf{91.30} & \textbf{90.02} & \textbf{91.14} & \textbf{88.78} & \textbf{95.18} & \textbf{91.54} & \textbf{83.91} & \textbf{88.68} & \textbf{86.14} & \textbf{87.25\(_{\pm0.06}\)} \\
\bottomrule
\end{tabular}%
}
\vspace{-15pt}
\label{tab:cifar10c}
\end{table*}
\subsection{Rectification}
\label{sec:method-rect}
To further refine the target model using the blended teacher, we construct a rectification matrix. 
By imposing marginal constraints on this matrix, we exert global control over within-batch class assignments, thereby preventing collapse or severe class imbalance (e.g., all samples being confidently mapped to a single class). 
Concretely, we introduce an optimal transport (OT)~\cite{ot1, ot2, silva2025conformal} step that reconciles the target model’s predictions within each mini-batch, yielding a rectified matrix \(p^{\text{rm}}_i\). 
Using \(p^{\text{rm}}_i\) as refined supervision improves the target model’s robustness at the mini-batch level. 
The rectification preserves fidelity to the original similarities \(p^{\text{tar}}_i\) while enforcing global margin constraints, resulting in smoothed, distribution-aligned soft scores. Formally, we define the following OT problem:
\begin{equation}
\label{eq:rec-1}
\max_{\mathcal{P}} \operatorname{tr}(\mathbf{P}^{\text{rm}} {}^{\top} \mathbf{P}^{\text{tar}}),
\end{equation}
where \(\mathbf{P}^{\text{tar}}=(p^{\text{tar}}_1, p^{\text{tar}}_2, \dots, p^{\text{tar}}_N)\in \mathbb{R}^{K\times N}\) with \(N\) the batch size. The transport plan \(\mathbf{P}^{\text{rm}}\in \mathbb{R}^{K\times N}\) is treated as the rectified distribution, i.e., \(p^{\text{rm}}_i=\mathbf{P}^{\text{rm}}_i\).
The plan must satisfy the marginal constraints:
\begin{equation}
\label{eq:rec-2}
\mathcal{P}=\{\mathbf{P}^{\text{rm}}\mid \mathbf{P}^{\text{rm}}\mathbf{1}_{N}=\mathbf{m},\ \mathbf{P}^{\text{rm}}{}^{\top}\mathbf{1}_{K}=\mathbf{u}_{N}\},
\end{equation}
where \(\mathbf{m}\) denotes the mini-batch label marginal.
We instantiate \(\mathbf{m}\) via pseudo-label voting over the three predictions, namely \(p^{\text{bt}}_i\), \(p^{\text{tar}}_i\), and \(p^{\text{tea}}_i\), which serves as a reliable proxy for the true label distribution. 
We relax the problem and solve it efficiently with the Sinkhorn algorithm~\cite{ot1}, which converges to \(\mathbf{P}^{\text{rm}}\) within a few iterations.

Finally, we refine the target model using mutual information between \(p^{\text{tar}}_i\) and \(p^{\text{rm}}_i\):
\begin{equation}
\mathcal{L}_{\text{Rect}}(x_i) = -\text{MI}(p^{\text{tar}}_i;p^{\text{rm}}_i),
\end{equation}
where \(\text{MI}(\cdot;\cdot)\) denotes the mutual information loss~\cite{ji2019invariant}.

\subsection{Overall Procedure of CoDiRe}
\label{sec:method-overall}
In Section~\ref{sec:method-distill} and ~\ref{sec:method-rect}, we introduced \(\mathcal{L}_{\text{Distill}}\) and \(\mathcal{L}_{\text{Rect}}\). 
As most TTA works do \cite{EATA,cliff,DEYO}, we further incorporate a widely adopted entropy loss to enable the target model to update with the distribution of the current data stream:
\begin{equation}
\mathcal{L}_{\text{Ent}}(x_i) = \frac{\mathcal{E}_i}{\exp(\mathcal{E}_i - \tau_{\text{Ent}})},
\end{equation}
where \(\mathcal{E}_i = -\sum_{c=1}^{K} p_{ic}^{\text{tar}} \log p_{ic}^{\text{tar}}\), and $\tau_{\text{ENT}}$ controls sensitivity to entropy. 
Therefore, the final loss function is:
\begin{equation}
\mathcal{L}_{\text{total}}=\mathcal{L}_{\text{Ent}}+\mathcal{L}_{\text{Distill}}+\mathcal{L}_{\text{Rect}}.
\end{equation}

Beyond the distillation mechanisms, we further address catastrophic forgetting in CTTA by endowing the target model with distribution-awareness. 
Prior work has explored reset heuristics~\cite{cotta, vida}; for example, CoTTA randomly resets a subset of parameters. 
In contrast, we neither reset all parameters nor resort to random resets. 
Instead, we design a distribution-aware layer-wise reset mechanism. 
Let the target model have parameters \(\theta_{t}\) at step \(t\). 
Specifically, we define an anchor \(\theta^{\text{anchor}}\), initialized as \(\theta_0\) and updated every \(s\) steps. We then introduce two displacement vectors \cite{domaindetector}:
\begin{align}
\delta_t &= \theta_{t} - \theta_{t-1}, \\
\delta^{\text{anchor}}_t &= \theta_{t-1} - \theta^{\text{anchor}}.
\end{align}

A divergence between \(\delta_t\) and \(\delta_t^{\text{anchor}}\) indicates a domain change. We quantify this via cosine similarity:
\begin{equation}
\gamma=\cos(\delta_t, \delta_t^{\text{anchor}}) = \frac{\langle \delta_t, \delta_t^{\text{anchor}} \rangle}{\|\delta_t\| \cdot \|\delta_t^{\text{anchor}}\|}.
\end{equation}
When \(\gamma<\gamma_0\), we regard this as a domain switch, with \(\gamma_0\) serving as the reset threshold. 
We observe that deeper layers tend to capture domain-specific activation statistics and are thus more susceptible to shift-induced drift, whereas shallower layers encode domain-invariant structural cues (e.g., shapes and edges). 
Accordingly, we selectively reset only the deep, domain-specific last $\alpha\%$ layers. 
Detailed sensitivity analyses for the reset step size \(s\), switch threshold \(\gamma_0\), reset ratio $\alpha$, and a more comprehensive discussion of the reset mechanism are provided in Section~\ref{sec:discussion}.

\begin{table*}[ht]
\caption{\textbf{Comparison results under corruption scenarios on ImageNet-C.} Classification accuracy of the standard ImageNet \(\rightarrow\) ImageNet-C online continual test-time adaptation task while continually adapting to different corruptions at the highest severity 5. The best and second-best results are highlighted in \textbf{bold} and \underline{underlined}, respectively.}
\vspace{-10pt}
\centering
\resizebox{0.95\textwidth}{!}{%
\begin{tabular}{c|c|ccccccccccccccc|c}
\toprule
\textbf{ImageNet-C} &
\multirow{2}{*}{\textbf{Venue}} &
\multicolumn{3}{c}{\textbf{Noise}} &
\multicolumn{4}{c}{\textbf{Blur}} &
\multicolumn{4}{c}{\textbf{Weather}} &
\multicolumn{4}{c|}{\textbf{Digital}} &
\multirow{2}{*}{\textbf{Avg.}} \\ %\cline{2-16}
\textbf{ViT-B/16} &  & \textit{gauss.} & \textit{shot} & \textit{impul.} & \textit{defoc.} & \textit{glass} & \textit{motion} & \textit{zoom} & \textit{snow} & \textit{frost} & \textit{fog} & \textit{brit.} & \textit{contr.} & \textit{elastic} & \textit{pixel} & \textit{jpeg} &  \\ 
\rowcolor{gray!10} & & \multicolumn{15}{c|}{\textit{\small{Time}}\hspace{2em}\textemdash\textemdash\textemdash\textemdash\textemdash\textemdash\textemdash\textemdash\textemdash\textemdash\textemdash\textemdash\textemdash\textemdash\textemdash\textemdash\textemdash\textemdash\textemdash\textemdash\textemdash\textemdash\textemdash\textemdash\textemdash\textemdash\textemdash\textemdash\textemdash\textemdash\textemdash\textemdash\textemdash\textemdash\textemdash\textemdash\textemdash\textemdash\textemdash\textemdash\textemdash\textemdash>} & \\ 
\midrule
\textbf{Source}
  & - & 35.06 & 33.68 & 36.92 & 32.46 & 23.04 & 36.78 & 30.56
  & 21.40 & 27.54 & 52.84 & 62.60 & 50.26 & 31.68 & 53.58 & 57.28 & 39.05 \\

\textbf{Tent}
  & ICLR'21 & 42.31 & 48.90 & 52.67 & 42.41 & 36.65 & 50.99 & 44.75
  & 48.97 & 54.74 & {65.37} & 74.75 & \textbf{61.08} & 45.43 & 64.92 & 66.70 & 53.38\(_{\pm0.10}\) \\

\textbf{MEMO}
  & NIPS'22 & 41.24 & 40.80 & 42.56 & 32.78 & 29.32 & 44.86 & 37.74 & 30.24 & 33.82 & 53.86 & 69.34 & 56.34 & 33.32 & 62.32 & 59.94 & 44.57\(_{\pm0.18}\) \\

\textbf{EATA}
  & ICML'22 & 47.93 & 55.20 & {57.23} & {49.21} & \underline{49.27} & {55.49} & \underline{53.18}
  & {59.45} & 62.47 & 65.27 & {76.71} & 57.82 & \textbf{57.09} & {67.63} & 67.07 & {58.73}\(_{\pm0.19}\) \\

\textbf{SAR}
  & ICLR'23 & 41.81 & 48.97 & 53.16 & 43.69 & 38.15 & 51.29 & 45.08
  & 48.23 & 54.93 & 64.56 & 75.48 & \underline{59.76} & 45.85 & 64.37 & 66.54 & 53.46\(_{\pm0.08}\) \\

\textbf{DeYO}
  & ICLR'24 & 50.18 & 56.75 & 57.95 & 48.67 & 49.84 & 54.75 & 48.15
  & 58.55 & 60.65 & 63.97 & 75.77 & {56.45} & {56.45} & 66.69 & {67.55} & 58.16\(_{\pm0.24}\) \\

\midrule
\textbf{CoTTA}
  & CVPR'22 & 54.37 & 53.67 & 54.89 & 50.02 & 32.02 & 52.47 & 45.30 & 59.83 & 62.41 & 64.37 & 77.76 & 34.84 & 45.33 & 66.67 & 69.53 & 54.90\(_{\pm0.06}\) \\

\textbf{NOTE}
  & NIPS'22 & 54.05 & 53.45 & 54.11 & 50.15 & 33.22 & 52.25  & 45.08 & 60.27 & 62.49 & 65.63  & \textbf{77.77} & 36.30 & 44.95 & 67.80  & 68.97 & 55.10\(_{\pm0.22}\) \\ 

\textbf{RoTTA}
  & CVPR'23 & 53.98 & 54.02 & 55.31 & 49.19 & 35.31 & 54.69 & 48.45 & \underline{62.87} & \textbf{65.08} & 64.54 & \textbf{77.77} & 38.08 & 50.17 & 68.35 & \underline{70.29} & 56.54\(_{\pm0.40}\) \\

\textbf{SANTA}
  & TMLR'23 & \textbf{55.99} & \textbf{59.27} & \textbf{59.69} & \underline{51.08} & 40.74 & \underline{56.63} & 50.87 & 62.55 & \underline{64.69} & \textbf{68.07} & 77.48 & 59.13 & 50.53 & 68.29 & 69.36 & \underline{59.62\(_{\pm0.62}\)} \\

\textbf{ViDA}
  & ICLR'24 & \underline{54.55} & 56.05 & \underline{57.65} & 50.18 & 34.78 & 54.69 & 46.92 & 61.55 & 63.25 & 64.61 & \underline{77.70} & 36.79 & 49.81 & \underline{69.26} & \textbf{70.49} & 56.55\(_{\pm0.61}\) \\
  
\textbf{DPCore}
  & ICML'25 & 51.15 & 53.25 & 53.25 & 40.06 & 41.36 & 53.95 & 47.21 & 58.91 & 55.35 & 53.97 & 75.37 & 51.49 & 51.39 & 67.35 & 63.54 & 54.51\(_{\pm1.02}\) \\

\midrule
\textbf{CLIP}
  & ICML'21 & 22.94 & 23.20 & 24.06 & 31.50 & 19.80 & 35.84 & 33.58
  & 45.00 & 39.34 & 47.26 & 62.88 & 34.54 & 25.74 & 50.68 & 42.02 & 35.89 \\

\rowcolor{cyan!10}
\textbf{Ours}
  & - & 54.43 & \underline{57.64} & 57.43 & \textbf{53.09} & \textbf{50.11} & \textbf{57.51} & \textbf{55.54} & \textbf{62.91} & 62.76 & \underline{66.73} & 77.57 & 58.65 & \underline{56.03} & \textbf{70.02} & 69.91 & \textbf{60.69\(_{\pm0.19}\)} \\
\bottomrule
\end{tabular}%
}
\vspace{-10pt}
\label{tab:imagenet}
\end{table*}
\section{Experiments}
\label{sec:exp}
\subsection{Experimental Setup}
\label{sec:setup}
Our experiments are designed to address the following key questions:
\textbf{RQ1:} How does CoDiRe perform in comparison to existing methods across various real-world scenarios?
\textbf{RQ2:} What is the contribution of each component of CoDiRe to its overall performance?
\textbf{RQ3:} How do these components and hyper-parameters work and what are their advantages over other alternatives or baselines?

To address the above questions, we evaluate CoDiRe in two distinct real-world scenarios: corruptions and domain generalizations.
As discussed before, all experiments are conducted under the standard CTTA protocols. 
In the corruption scenarios, we conduct evaluations on CIFAR-10-C and ImageNet-C \cite{cifar10-100-C}, across 15 different corruption types sequentially at the highest severity level 5. 
In the domain generalization scenarios, we use two widely adopted datasets: OfficeHome \cite{officehome} and PACS \cite{PACS}. 
Here, we treat one domain as the source domain and concatenate the remaining domains sequentially for continual adaptation. 
Across both settings, we compare CoDiRe against a comprehensive set of baselines, including:
1) TTA methods: BN Adapt \cite{bn_adapt}, Tent \cite{Tent}, MEMO \cite{Memo}, EATA \cite{EATA}, SAR \cite{SAR}, and DeYO \cite{DEYO}; 
2) CTTA methods: CoTTA \cite{cotta}, NOTE \cite{note}, RoTTA \cite{rotta}, SANTA \cite{santa}, ViDA \cite{vida}, and DPCore \cite{dpcore}.  

Note that our method's use of CLIP is similar to another popular line of work, namely VLM-TTA, which aims to adapt CLIP itself on OOD data during test time.
Though the objective and task differ from ours, we include these methods for a comprehensive and fair comparison: 3) VLM-TTA methods: TPT \cite{tpt}, TDA \cite{tda}, BoostAdapter \cite{boostadapter}, and ZERO \cite{zero}.
Since there are no existing baselines under our proposed TTD paradigm, we also design several 4) TTD methods: Naive Ensemble (NE), BN Adapt w. NE, Tent w. NE, and Distill. CLIP, as detailed in Section \ref{sec:exp-main}.

In implementation, we consistently adopt CLIP ViT-L/14 as the VLM teacher \(\mathcal{F}(\cdot)\). 
For the target model \(f(\cdot)\), we employ a ViT-B/16~\cite{vit} backbone architecture on ImageNet-C, and ResNet-50~\cite{resnet} on the other three datasets following prior work~\cite{DEYO}.
For more information about experimental details, please refer to the Supplementary Materials. 

\subsection{Main Results}
\label{sec:exp-main}
\paragraph{Comparisons on Corruptions Scenarios.}
We first report the comparisons on the two corruption datasets CIFAR-10-C and ImageNet-C, as shown in Table \ref{tab:cifar10c} and \ref{tab:imagenet}, respectively. 
As noted in Section~\ref{sec:empirical}, under domain shift, CLIP underperforms a supervised classifier with the same backbone. Even CLIP ViT-L/14 lags behind a smaller ViT-B/16 trained on ImageNet in terms of performance on ImageNet-C.

In contrast, CoDiRe, by constructing a more reliable blended teacher as the distillation target, achieves the best overall performance across both datasets. On CIFAR-10-C and ImageNet-C, it not only surpasses CLIP’s zero-shot capability but also outperforms state-of-the-art CTTA methods, achieving 87.25\% and 60.69\%, respectively. Notably, CoDiRe keeps CLIP frozen, without requiring access to its parameters, gradients, or architectural details. It can even operate via a simple VLM API, rendering it highly practical for real-world deployment.

\paragraph{Comparisons on Domain Generalization Scenarios.}
\begin{table*}[t]
\captionof{table}{\textbf{Comparison results under domain generalization scenarios on OfficeHome and PACS.} Classification accuracy of the standard OfficeHome-Art/PACS-Art \(\rightarrow\) the rest domains online continual test-time adaptation task while continually adapting to different domains. The best and second-best results are highlighted in \textbf{bold} and \underline{underlined}, respectively.}
\label{tab:dg}
\vspace{-10pt}
\resizebox{\textwidth}{!}{
% \begin{tabular}{*{16}{c}}
\begin{tabular}{cc|ccccccc|ccccc|cc}
\specialrule{1.2pt}{1pt}{1pt} % Top thick line
% \hline
\multirow{2}{*}{\textbf{Source}} & \multirow{2}{*}{\textbf{Target}} & \textbf{Source} & \textbf{BN Adapt} & \textbf{Tent} & \textbf{MEMO} & \textbf{EATA} & \textbf{SAR} & \textbf{DeYO} & \textbf{CoTTA} & \textbf{NOTE} & \textbf{RoTTA} & \textbf{SANTA} & \textbf{ViDA} & \textbf{CLIP} & \cellcolor{cyan!10}\textbf{Ours} \\ 
& & --- & NIPS'20 & ICLR'21 & NIPS'22 & ICML'22 & ICLR'23 & ICLR'24 & CVPR'22 & NIPS'22 & CVPR'23 & TMLR'23 & ICLR'24 & ICML'21 & \cellcolor{cyan!10}- \\ 
\hline
\multirow{4}{*}{\makecell{\textbf{OfficeHome}\\\textit{Art}}} & \textit{Clipart} & 47.93 & 46.97 & 47.36 & 48.24 & 47.51 & 47.29 & 48.17 & 40.84 & 49.47 & 48.94 & 48.37 & 48.19 & \underline{67.67} & \cellcolor{cyan!10}\textbf{70.98} \\

& \textit{Product} & 65.78 & 61.05 & 61.25 & 65.01 & 61.28 & 61.12 & 60.27 & 44.01 & 64.59 & 63.37 & 62.70 & 62.53 & \underline{84.73} & \cellcolor{cyan!10}\textbf{85.39} \\

& \textit{Real} & 73.24 & 70.13 & 69.76 & 72.76 & 69.83 & 69.80 & 67.95 & 60.76 & 71.10 & 71.99 & 71.18 & 71.29 & \underline{83.70} & \cellcolor{cyan!10}\textbf{84.77} \\

& \textbf{Avg.} & 62.32 & 59.39\(_{\pm0.29}\) & 59.46\(_{\pm0.18}\) & 62.00\(_{\pm0.04}\) & 59.54\(_{\pm0.18}\) & 59.40\(_{\pm0.22}\) & 58.80\(_{\pm0.13}\) & 48.54\(_{\pm0.13}\) & 61.72\(_{\pm0.03}\) & 61.43\(_{\pm0.14}\) & 60.75\(_{\pm0.17}\) & 60.67\(_{\pm0.10}\) & \underline{78.70} & \cellcolor{cyan!10}\textbf{80.38}\(_{\pm0.04}\) \\
\hline
\multirow{4}{*}{\makecell{\textbf{PACS}\\ \textit{Art}}} & \textit{Cartoon} & 66.04 & 74.80 & 74.97 & 69.78 & 74.80 & 74.93 & 75.68 & 75.41 & 68.87 & 73.04 & 74.97 & 74.79 & \underline{99.53} & \cellcolor{cyan!10}\textbf{99.73} \\

& \textit{Photo} & 97.84 & 96.83 & 96.89 & 98.20 & 96.83 & 96.79 & 97.09 & 97.07 & 97.45 & 93.73 & 96.89 & 96.81 & \textbf{99.88} & \cellcolor{cyan!10}\underline{99.84} \\

& \textit{Sketch} & 57.32 & 69.32 & 70.73 & 60.84 & 69.32 & 69.72 & 71.67 & 74.15 & 65.65 & 71.80 & 71.43 & 69.48 & \underline{95.24} & \cellcolor{cyan!10}\textbf{95.41} \\

& \textbf{Avg.} & 73.73 & 80.32\(_{\pm0.18}\) & 80.86\(_{\pm0.11}\) & 76.27\(_{\pm0.08}\) & 80.32\(_{\pm0.18}\) & 80.48\(_{\pm0.15}\) & 81.48\(_{\pm0.25}\) & 82.21\(_{\pm0.16}\) & 77.32\(_{\pm0.11}\) & 79.52\(_{\pm0.10}\) & 81.10\(_{\pm0.15}\) & 80.36\(_{\pm0.18}\) & \underline{98.22} & \cellcolor{cyan!10}\textbf{98.33}\(_{\pm0.01}\) \\

% \hline
\specialrule{1.2pt}{1pt}{1pt} % Top thick line
\end{tabular}
}
\vspace{-17pt}
\end{table*}
\begin{table}
\caption{\textbf{Comparison results with VLM-TTA and TTD baselines.} Average classification accuracy of the standard CTTA task on CIFAR-10-C, ImageNet-C, OfficeHome and PACS, with the same setting as Table~\ref{tab:cifar10c}, Table~\ref{tab:imagenet} and Table~\ref{tab:dg}. The best and second-best results are highlighted in \textbf{bold} and \underline{underlined}, respectively.}
\vspace{-8pt}
\label{table:vlmtta}
\resizebox{\linewidth}{!}{
\begin{tabular}{c|c|cccc|c}
\specialrule{1.2pt}{1pt}{1pt} % Top thick line
% \hline
\textbf{Method} & \textbf{Venue} & \textbf{CIFAR-10-C} & \textbf{ImageNet-C} & \textbf{OfficeHome} & \textbf{PACS} & \textbf{Avg.} \\
\hline
\textbf{CLIP} & ICML' 21 & 74.39 & 35.89 & 78.70 & 98.22 & 71.80 \\
\hline
\textbf{TPT} & NIPS' 22 & 73.52\(_{\pm0.02}\) & 36.24\(_{\pm0.05}\) & 79.07\(_{\pm0.01}\) & \underline{98.32}\(_{\pm0.01}\) & 71.79 \\

\textbf{TDA} & CVPR' 24 & 75.61\(_{\pm0.15}\) & 37.55\(_{\pm0.32}\) & 79.75\(_{\pm0.21}\) & 98.13\(_{\pm0.07}\) & 72.76 \\

\textbf{BoostAdapter} & NIPS' 24 & 75.80\(_{\pm0.17}\) & 38.14\(_{\pm0.17}\) & \underline{80.25}\(_{\pm0.12}\) & 98.18\(_{\pm0.06}\) & 73.09 \\

\textbf{ZERO} & NIPS' 24 & 77.89\(_{\pm0.04}\) & 17.43\(_{\pm0.17}\) & 79.46\(_{\pm0.10}\) & 97.66\(_{\pm0.05}\) & 68.11 \\

\hline
\textbf{Naive Ensemble} & - & 76.90\(_{\pm0.00}\) & 47.95\(_{\pm0.27}\) & 79.89\(_{\pm0.00}\) & 96.29\(_{\pm0.00}\) & 75.26 \\

\textbf{BN Adapt w. NE} & - & 84.56\(_{\pm0.02}\) & 47.95\(_{\pm0.27}\) & 80.23\(_{\pm0.08}\) & 97.55\(_{\pm0.06}\) & 77.57 \\

\textbf{Tent w. NE} & - & \underline{86.29}\(_{\pm0.03}\) & \underline{56.58}\(_{\pm0.24}\) & \underline{80.25}\(_{\pm0.08}\) & 97.57\(_{\pm0.08}\) & 80.17 \\

\textbf{Distil. CLIP} & - & 77.52\(_{\pm0.05}\) & 48.29\(_{\pm0.13}\) & 61.02\(_{\pm0.12}\) & 78.34\(_{\pm0.35}\) & 66.29 \\

\rowcolor{cyan!10}
\textbf{Ours} & - & \textbf{87.25}\(_{\pm0.06}\) & \textbf{60.69}\(_{\pm0.19}\) & \textbf{80.38}\(_{\pm0.07}\) & \textbf{98.33}\(_{\pm0.06}\) & \textbf{81.66} \\

% \hline
\specialrule{1.2pt}{1pt}{1pt} % Top thick line
\end{tabular}
\vspace{-10pt}
}
\end{table}
We next evaluate CoDiRe in domain generalization scenarios on OfficeHome and PACS, as shown in Table \ref{tab:dg}. 
Unlike corruption scenarios where images are severely degraded, shifts here arise from stylistic or contextual variations across domains (e.g., cartoons and sketches). 
Thus, CLIP exhibits robust performance in such scenarios, which greatly benefits from large-scale pre-training on diverse web imagery.
Consistent with this, only a zero-shot CLIP could significantly outperform all the TTA/CTTA baselines. 
Nonetheless, CoDiRe still establishes excellent performance across both datasets. 
These results highlight CoDiRe's ability to balance the complementary strengths of the task-specific knowledge from the target model and the open-world knowledge from CLIP. 
It remains effective across diverse shift types, from low-level corruptions to high-level domain variation.

\paragraph{Comparisons with VLM-TTA and TTD baselines.}
A substantial body of work focuses on directly enhancing CLIP’s generalization ability during inference, collectively referred to as VLM-TTA. However, recent evidence~\cite{maharana2025batclip} indicates that these methods perform poorly on corruption benchmarks such as CIFAR-10-C and ImageNet-C. 
In contrast, CoDiRe, equipped with the target model, still achieves state-of-the-art performance on both corruption and domain generalization datasets, as shown in Table~\ref{table:vlmtta}. 
This demonstrates that CoDiRe not only improves the target model, but also benefits CLIP itself at test-time.

Moreover, we evaluate several naive TTD baselines. 
A straightforward average interpolation yields a simple prediction \(p_i^{\text{ens}}\), where \(p_i^{\text{ens}}=\sigma(\mathbf{z}_i^{\text{esn}}),\  \mathbf{z}_i^{\text{esn}}=\frac{1}{2}(\mathbf{z}_i^{\text{tar}}+\mathbf{z}_i^{\text{tea}})\), which we term Naive Ensemble (NE). 
We further compare BN Adapt and Tent with NE, and evaluate the performance when directly using CLIP logits as the distillation target. 
The results in Table~\ref{table:vlmtta} show that CoDiRe surpasses these naive approaches by constructing a more effective blended teacher, thereby achieving the best performance. 
The poor results of Distill. CLIP further validates our Generalist Trap pitfall.

\begin{table}[htbp]
    \caption{\textbf{Ablation study on two corruption datasets.} Average classification accuracy of the standard CTTA task on CIFAR-10-C and ImageNet-C, with the same setting as Table~\ref{tab:cifar10c} and Table~\ref{tab:imagenet}. The best and second-best results are highlighted in \textbf{bold} and \underline{underlined}.}
    \vspace{-20pt}
    \label{tab:component}
    \begin{center}
    \begin{threeparttable}
    % \LARGE % Set font size for the table
    \resizebox{\linewidth}{!}{ % Automatically scale table to page width
    \begin{tabular}{l|cccc|cc|c}
    \specialrule{1.2pt}{1pt}{1pt} % Top thick line
      & \multicolumn{4}{c|}{Components} & \multirow{2}{*}{CIFAR-10-C} & \multirow{2}{*}{ImageNet-C} & \multirow{2}{*}{Avg.} \\
    % \cmidrule(lr){2-4} \cmidrule(lr){5-7} \cmidrule(lr){8-10}
     & \(\mathcal{L}_\text{Distill}\) & \(\mathcal{L}_\text{Rect}\) & \(\mathcal{L}_\text{Ent}\) & reset &  &  &  \\
    \midrule
    (1) BT &   &  &  &  & 84.46\(_{\pm0.02}\) & 48.05\(_{\pm0.15}\) & 66.26 \\
    (2)  &   & \ding{51} & \ding{51} & \ding{51} & 86.71\(_{\pm0.02}\) & 59.30\(_{\pm0.05}\) & 73.01 \\
    (3)  & \ding{51}  &  &  & \ding{51} & 87.02\(_{\pm0.07}\) & 59.82\(_{\pm0.21}\) & 73.42 \\
    (4)  &  \ding{51} & \ding{51} &  & \ding{51} & \underline{87.12}\(_{\pm0.06}\) & 60.30\(_{\pm0.14}\) & 73.71 \\
    (5)  &  \ding{51} &  &  \ding{51}  & \ding{51} & 87.11\(_{\pm0.08}\) & 60.15\(_{\pm0.12}\) & 73.63 \\
    (6)  & \ding{51} & \ding{51} & \ding{51} &  & 86.71\(_{\pm0.08}\) & \underline{60.41}\(_{\pm0.86}\) & 73.56 \\
    \rowcolor{cyan!10}(7) Ours  & \ding{51} & \ding{51} & \ding{51} & \ding{51} & \textbf{87.25}\(_{\pm0.06}\) & \textbf{60.69}\(_{\pm0.19}\) & \textbf{73.97} \\
    \specialrule{1.2pt}{1pt}{1pt} % Bottom thick line
    \end{tabular}
    % \vspace{0.5cm}
    }
    \end{threeparttable}
    \end{center}
    \vspace{-10pt}
\end{table}

\subsection{Ablation Study}
\label{sec:ablation}
We conduct an ablation study to assess the contributions of the three loss terms and the distribution-aware reset mechanism. 
Table \ref{tab:component} reports the results on two corruption benchmarks. 
Notably, even without any gradient backpropagation, the blended teacher already outperforms both the target model and CLIP.
Building on this, incorporating each component yields additional gains: $\mathcal{L}_{\text{Ent}}$ enhances prediction confidence, 
$\mathcal{L}_{\text{Distill}}$ distills the rich knowledge from blended teacher, 
$\mathcal{L}_{\text{Rect}}$ further rectifies the target model, 
and our reset mechanism alleviates catastrophic forgetting during CTTA.

\subsection{Discussions}
\label{sec:discussion}
\paragraph{Hyper-parameters.}
\begin{figure}
    \vspace{-10pt} 
    \centering
    \includegraphics[width=0.99\linewidth]{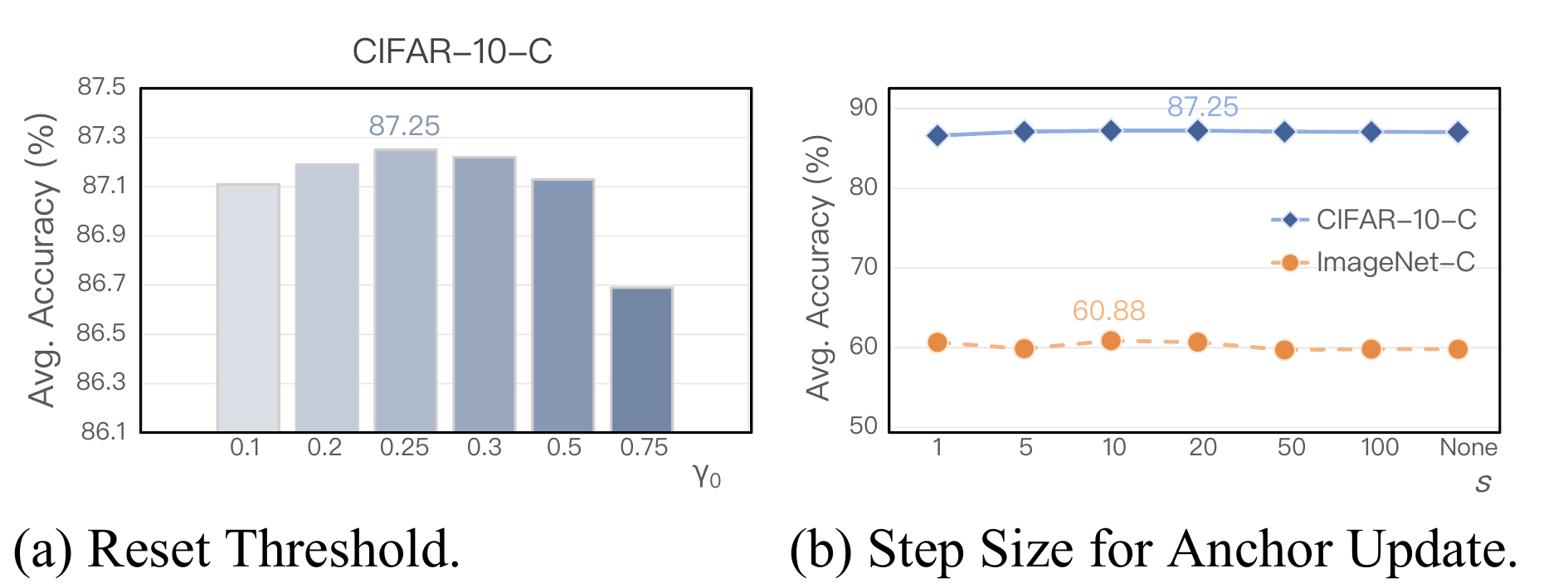}
    \captionsetup{width=0.99\linewidth}  % Set caption width to match image width
    \vspace{-10pt}
    \caption{
    \textbf{Hyper-parameters of Reset Mechanisms.} (a) Hyperparameter study of the reset threshold \(\gamma_0\) on CIFAR-10-C. (b) Hyperparameter study of the step size \(s\) for updating the anchor \(\theta^{\text{anchor}}\) on CIFAR-10-C and ImageNet-C.
}
    \label{fig:hyper}
    \vspace{-0.5cm} 
\end{figure}
\par  % End the paragraph and reset wrapping
\noindent  % Prevent indentation of the next line
We conduct a sensitivity experiment on the hyper-parameters \(\gamma_0\) and \(s\) introduced in our reset mechanisms. 
As illustrated in Figure~\ref{fig:hyper}, CoDiRe exhibits strong robustness to both hyper-parameters: performance remains stable across wide intervals, with only minor fluctuations at extreme values. 

\begin{figure}
    \vspace{-15pt} 
    \centering
    \includegraphics[width=0.99\linewidth]{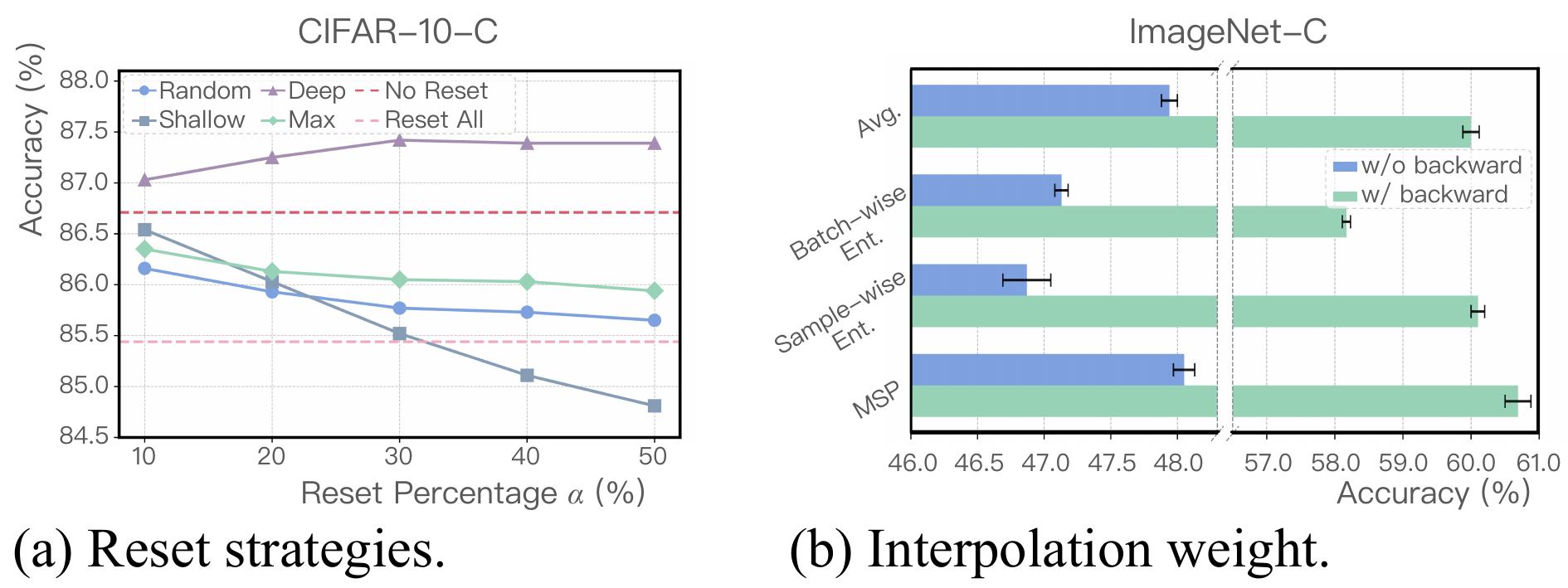}
    \captionsetup{width=0.99\linewidth}  % Set caption width to match image width
    \vspace{-10pt}
    \caption{
    \textbf{Discussions of CoDiRe.} (a) Effect of reset strategies under varying percentages \(\alpha\). (b) Effect of different fusion weights.
}
    \label{fig:exp_fusion}
    \vspace{-0.5cm} 
\end{figure}

\paragraph{Reset Strategies.}
We also investigate various reset strategies for the target model, including resetting shallow layers, deep layers, randomly selected layers, and layers with maximum drift, along with no reset and full reset as baselines. 
As shown in Figure \ref{fig:exp_fusion}(a), selectively resetting deep layers consistently delivers the strongest performance across a wide range of reset percentages $\alpha$. 
This corroborates our prior observation that deep layers are more prone to accumulating detrimental domain-specific drift due to their role in modeling higher-level activation statistics and semantic features. 
In contrast, most alternatives underperform even compared with no reset, as they disrupt beneficial adaptation or interfere with learning of domain-invariant features. 
Resetting based on maximum shifts is also suboptimal, likely due to heterogeneous learning dynamics and feature distributions across layers. 
Overall, these results underscore the importance of strategic, layer-aware resetting for robust CTTA.

\paragraph{Fusion Weights.}
We explore different fusion schemes \(\lambda\) in Equation \ref{eq:fusion} to combine the logits of the two models. 
Beyond simple averaging, we adopt dynamic weighting based on batch-wise and sample-wise entropy, alongside our proposed MSP-based approach. 
Figure \ref{fig:exp_fusion}(b) reports results on ImageNet-C. 
Direct averaging is a surprisingly strong and straightforward baseline; however, our dynamic MSP-based weighting is overall more robust than fixed averaging. In contrast, entropy-based weights exhibit limited effectiveness and can sometimes cause severe performance degradations, consistent with our Entropy Bias pitfall.

\par  % End the paragraph and reset wrapping
\noindent  % Prevent indentation of the next line

\paragraph{Efficiency Analysis.}
\begin{table}[htbp]
\centering
\vspace{-20pt}
\caption{\textbf{Efficiency analysis on ImageNet-C.} All methods are evaluated on a single NVIDIA Tesla V100 GPU.}
\vspace{-10pt}
\label{tab:efficiency}
\resizebox{\linewidth}{!}{ % Resize table to fit page width
\begin{tabular}{l|cccccc} 
\toprule
\textbf{Method} & \textbf{Source} & \textbf{Tent} & \textbf{CoTTA} & \textbf{CLIP} &  \textbf{Ours} \\ 
\midrule
\textbf{Memory (GiB) $\downarrow$} & 
1.68 & 6.62 & 18.41 & 4.24 & 8.18 \\

\textbf{Testing Time (min) $\downarrow$} & 
4.23 & 13.43 & 36.62 & 10.14 & 17.86 \\

\rowcolor{cyan!10}\textbf{Avg Acc. $\uparrow$} & 
39.05 & 53.38 & 54.90 & 35.89 & \textbf{60.69}  \\
\bottomrule
\end{tabular}
\vspace{-5pt}
}
\end{table}

Introducing CLIP as the teacher model inevitably incurs additional overhead. 
To quantify this, we report the time and memory consumption on ImageNet-C in Table~\ref{tab:efficiency}. 
Since CoDiRe keeps CLIP frozen, its extra memory overhead is less than only 15\% of Tent's; and the computational cost amounts to only one additional CLIP forward pass, increasing overall latency by merely about 30\%. 
In contrast, CoTTA retains the source, teacher and student model simultaneously, and applies computationally heavy augmentations to maintain robustness, which harms both space and time efficiency.
These results also suggest that TTD is a promising, efficient research paradigm.

\section{Conclusion}
In this work, we proposed Test-Time Distillation (TTD), a new paradigm for Continual Test-Time Adaptation (CTTA) that utilizes a frozen Vision-Language Model as an external signal to mitigate the error accumulation common in self-supervised methods. 
From comprehensive empirical studies, we identify two critical pitfalls in the practice of TTD: Generalist Trap and Entropy Bias. 
Our proposed method, CoDiRe, addresses these pitfalls by first constructing a robust blended teacher using a more reliable confidence metric based on Maximum Softmax Probability for distillation. 
It then further refines the target model's predictions with an Optimal Transport-based rectification step guided by the blended teacher. 
Extensive experiments show that CoDiRe achieves SoTA performance on standard CTTA benchmarks.

\newpage
\section*{Acknowledgment}
This work was supported by the National Natural Science Foundation of China (Grant No. 92467204 and 62472249), the Shenzhen Science and Technology Program (Grant No. KJZD20240903102300001 and JCYJ20250604145014018), and the Natural Science Foundation of Top Talent of SZTU (Grant No. GDRC202413). We thank the anonymous reviewers for their efforts, which have helped improve the quality of this paper.

{
    \small
    \bibliographystyle{ieeenat_fullname}
    \bibliography{main}

\begin{thebibliography}{78}
\providecommand{\natexlab}[1]{#1}
\providecommand{\url}[1]{\texttt{#1}}
\expandafter\ifx\csname urlstyle\endcsname\relax
  \providecommand{\doi}[1]{doi: #1}\else
  \providecommand{\doi}{doi: \begingroup \urlstyle{rm}\Url}\fi

\bibitem[Achiam et~al.(2023)Achiam, Adler, Agarwal, Ahmad, Akkaya, Aleman, Almeida, Altenschmidt, Altman, Anadkat, et~al.]{achiam2023gpt}
Josh Achiam, Steven Adler, Sandhini Agarwal, Lama Ahmad, Ilge Akkaya, Florencia~Leoni Aleman, Diogo Almeida, Janko Altenschmidt, Sam Altman, Shyamal Anadkat, et~al.
\newblock Gpt-4 technical report.
\newblock \emph{arXiv preprint arXiv:2303.08774}, 2023.

\bibitem[Anderson et~al.(2018)Anderson, He, Buehler, Teney, Johnson, Gould, and Zhang]{anderson2018bottom}
Peter Anderson, Xiaodong He, Chris Buehler, Damien Teney, Mark Johnson, Stephen Gould, and Lei Zhang.
\newblock Bottom-up and top-down attention for image captioning and visual question answering.
\newblock In \emph{Proceedings of the IEEE conference on computer vision and pattern recognition}, pages 6077--6086, 2018.

\bibitem[Chakrabarty et~al.(2023)Chakrabarty, Sreenivas, and Biswas]{santa}
Goirik Chakrabarty, Manogna Sreenivas, and Soma Biswas.
\newblock {SANTA}: Source anchoring network and target alignment for continual test time adaptation.
\newblock \emph{Transactions on Machine Learning Research}, 2023.

\bibitem[Chen et~al.(2023)Chen, Xie, Liu, Wang, Luo, Wang, and Zheng]{chen2023easynet}
Ruitao Chen, Guoyang Xie, Jiaqi Liu, Jinbao Wang, Ziqi Luo, Jinfan Wang, and Feng Zheng.
\newblock Easynet: An easy network for 3d industrial anomaly detection.
\newblock In \emph{Proceedings of the 31st ACM International Conference on Multimedia}, pages 7038--7046, 2023.

\bibitem[Chen et~al.(2024)Chen, Zhang, and Wang]{cliff}
Xiao Chen, Qihui Zhang, and Yan Wang.
\newblock Cliff: Leveraging ambiguous samples for enhanced test-time adaptation.
\newblock In \emph{27th European Conference on Artificial Intelligence}, pages 642--649, 2024.

\bibitem[Chen et~al.(2025)Chen, Du, Huang, Jiang, Lu, Jiang, and Wang]{chen2025neural}
Xiao Chen, Zhongjing Du, Jiazhen Huang, Xu Jiang, Li Lu, Jingyan Jiang, and Zhi Wang.
\newblock Neural collapse in test-time adaptation.
\newblock \emph{arXiv preprint arXiv:2512.10421}, 2025.

\bibitem[Cherti et~al.(2023)Cherti, Beaumont, Wightman, Wortsman, Ilharco, Gordon, Schuhmann, Schmidt, and Jitsev]{openclip}
Mehdi Cherti, Romain Beaumont, Ross Wightman, Mitchell Wortsman, Gabriel Ilharco, Cade Gordon, Christoph Schuhmann, Ludwig Schmidt, and Jenia Jitsev.
\newblock Reproducible scaling laws for contrastive language-image learning.
\newblock In \emph{Proceedings of the IEEE/CVF conference on computer vision and pattern recognition}, pages 2818--2829, 2023.

\bibitem[Cuturi(2013)]{ot1}
Marco Cuturi.
\newblock Sinkhorn distances: Lightspeed computation of optimal transport.
\newblock \emph{Advances in neural information processing systems}, 26, 2013.

\bibitem[Dong et~al.(2020)Dong, Yu, Cao, Shi, and Ma]{ensemble2}
Xibin Dong, Zhiwen Yu, Wenming Cao, Yifan Shi, and Qianli Ma.
\newblock A survey on ensemble learning.
\newblock \emph{Frontiers of Computer Science}, 14\penalty0 (2):\penalty0 241--258, 2020.

\bibitem[Dosovitskiy et~al.(2021)Dosovitskiy, Beyer, Kolesnikov, Weissenborn, Zhai, Unterthiner, Dehghani, Minderer, Heigold, Gelly, Uszkoreit, and Houlsby]{vit}
Alexey Dosovitskiy, Lucas Beyer, Alexander Kolesnikov, Dirk Weissenborn, Xiaohua Zhai, Thomas Unterthiner, Mostafa Dehghani, Matthias Minderer, Georg Heigold, Sylvain Gelly, Jakob Uszkoreit, and Neil Houlsby.
\newblock An image is worth 16x16 words: Transformers for image recognition at scale.
\newblock \emph{ICLR}, 2021.

\bibitem[Fan et~al.(2025)Fan, Jiang, Chen, Huang, Chen, Jiang, Zhang, Tang, and Wang]{fan2025moetta}
Xiao Fan, Jingyan Jiang, Zhaoru Chen, Fanding Huang, Xiao Chen, Qinting Jiang, Bowen Zhang, Xing Tang, and Zhi Wang.
\newblock Moetta: Test-time adaptation under mixed distribution shifts with moe-layernorm.
\newblock \emph{arXiv preprint arXiv:2511.13760}, 2025.

\bibitem[Fang et~al.(2021)Fang, Wang, Hu, Wang, Yang, and Liu]{fang2021compressing}
Zhiyuan Fang, Jianfeng Wang, Xiaowei Hu, Lijuan Wang, Yezhou Yang, and Zicheng Liu.
\newblock Compressing visual-linguistic model via knowledge distillation.
\newblock In \emph{Proceedings of the IEEE/CVF International Conference on Computer Vision}, pages 1428--1438, 2021.

\bibitem[Farina et~al.(2024)Farina, Franchi, Iacca, Mancini, and Ricci]{zero}
Matteo Farina, Gianni Franchi, Giovanni Iacca, Massimiliano Mancini, and Elisa Ricci.
\newblock Frustratingly easy test-time adaptation of vision-language models.
\newblock \emph{Advances in Neural Information Processing Systems}, 37:\penalty0 129062--129093, 2024.

\bibitem[Ganaie et~al.(2022)Ganaie, Hu, Malik, Tanveer, and Suganthan]{ensemble1}
Mudasir~A Ganaie, Minghui Hu, Ashwani~Kumar Malik, Muhammad Tanveer, and Ponnuthurai~N Suganthan.
\newblock Ensemble deep learning: A review.
\newblock \emph{Engineering Applications of Artificial Intelligence}, 115:\penalty0 105151, 2022.

\bibitem[Gong et~al.(2022)Gong, Jeong, Kim, Kim, Shin, and Lee]{note}
Taesik Gong, Jongheon Jeong, Taewon Kim, Yewon Kim, Jinwoo Shin, and Sung-Ju Lee.
\newblock {NOTE}: Robust continual test-time adaptation against temporal correlation.
\newblock In \emph{Advances in Neural Information Processing Systems (NeurIPS)}, 2022.

\bibitem[He et~al.(2016)He, Zhang, Ren, and Sun]{resnet}
Kaiming He, Xiangyu Zhang, Shaoqing Ren, and Jian Sun.
\newblock Deep residual learning for image recognition.
\newblock In \emph{Proceedings of the IEEE conference on computer vision and pattern recognition}, pages 770--778, 2016.

\bibitem[Hendrycks and Dietterich(2019)]{cifar10-100-C}
Dan Hendrycks and Thomas Dietterich.
\newblock Benchmarking neural network robustness to common corruptions and perturbations.
\newblock \emph{Proceedings of the International Conference on Learning Representations}, 2019.

\bibitem[Huang et~al.(2025)Huang, Jiang, Jiang, Li, Khan, and Wang]{huang2025cosmic}
Fanding Huang, Jingyan Jiang, Qinting Jiang, Hebei Li, Faisal~Nadeem Khan, and Zhi Wang.
\newblock Cosmic: Clique-oriented semantic multi-space integration for robust clip test-time adaptation.
\newblock In \emph{Proceedings of the Computer Vision and Pattern Recognition Conference}, pages 9772--9781, 2025.

\bibitem[Huang et~al.(2024)Huang, Zeng, Yang, An, Diao, and Xu]{huang2024etag}
Libo Huang, Yan Zeng, Chuanguang Yang, Zhulin An, Boyu Diao, and Yongjun Xu.
\newblock etag: Class-incremental learning via embedding distillation and task-oriented generation.
\newblock In \emph{Proceedings of the AAAI Conference on Artificial Intelligence}, pages 12591--12599, 2024.

\bibitem[Ji et~al.(2019)Ji, Henriques, and Vedaldi]{ji2019invariant}
Xu Ji, Joao~F Henriques, and Andrea Vedaldi.
\newblock Invariant information clustering for unsupervised image classification and segmentation.
\newblock In \emph{Proceedings of the IEEE/CVF international conference on computer vision}, pages 9865--9874, 2019.

\bibitem[Jia et~al.(2021)Jia, Yang, Xia, Chen, Parekh, Pham, Le, Sung, Li, and Duerig]{vlm2}
Chao Jia, Yinfei Yang, Ye Xia, Yi-Ting Chen, Zarana Parekh, Hieu Pham, Quoc Le, Yun-Hsuan Sung, Zhen Li, and Tom Duerig.
\newblock Scaling up visual and vision-language representation learning with noisy text supervision.
\newblock In \emph{International conference on machine learning}, pages 4904--4916. PMLR, 2021.

\bibitem[Jiang et~al.(2024)Jiang, Zhou, Li, Zhao, Wang, Ma, Chang, Zhang, Lu, et~al.]{jiang2024pcotta}
Jincen Jiang, Qianyu Zhou, Yuhang Li, Xinkui Zhao, Meili Wang, Lizhuang Ma, Jian Chang, Jian Zhang, Xuequan Lu, et~al.
\newblock Pcotta: Continual test-time adaptation for multi-task point cloud understanding.
\newblock \emph{Advances in Neural Information Processing Systems}, 37:\penalty0 96229--96253, 2024.

\bibitem[Jiang et~al.(2025)Jiang, Ye, Wei, Wang, Xue, Jiang, and Wang]{jiang2025feature}
Qinting Jiang, Chuyang Ye, Dongyan Wei, Bingli Wang, Yuan Xue, Jingyan Jiang, and Zhi Wang.
\newblock Feature-based instance neighbor discovery: Advanced stable test-time adaptation in dynamic world.
\newblock \emph{arXiv preprint arXiv:2506.06782}, 2025.

\bibitem[Karmanov et~al.(2024)Karmanov, Guan, Lu, El~Saddik, and Xing]{tda}
Adilbek Karmanov, Dayan Guan, Shijian Lu, Abdulmotaleb El~Saddik, and Eric Xing.
\newblock Efficient test-time adaptation of vision-language models.
\newblock \emph{The IEEE/CVF Conference on Computer Vision and Pattern Recognition}, 2024.

\bibitem[Khan et~al.(2025)Khan, Dambandkhameneh, Shaikh, Nie, Venugopal, and Li]{khan2025slidemambaentropybasedadaptivefusion}
Shakib Khan, Fariba Dambandkhameneh, Nazim Shaikh, Yao Nie, Raghavan Venugopal, and Xiao Li.
\newblock Slidemamba: Entropy-based adaptive fusion of gnn and mamba for enhanced representation learning in digital pathology, 2025.

\bibitem[Kim et~al.(2024)Kim, Park, and Choo]{kim2024model}
Dongmin Kim, Sunghyun Park, and Jaegul Choo.
\newblock When model meets new normals: test-time adaptation for unsupervised time-series anomaly detection.
\newblock In \emph{Proceedings of the AAAI conference on artificial intelligence}, pages 13113--13121, 2024.

\bibitem[Kim et~al.(2018)Kim, Jun, and Zhang]{kim2018bilinear}
Jin-Hwa Kim, Jaehyun Jun, and Byoung-Tak Zhang.
\newblock Bilinear attention networks.
\newblock \emph{Advances in neural information processing systems}, 31, 2018.

\bibitem[Lee et~al.(2024)Lee, Jung, Lee, Park, Shin, Hwang, and Yoon]{DEYO}
Jonghyun Lee, Dahuin Jung, Saehyung Lee, Junsung Park, Juhyeon Shin, Uiwon Hwang, and Sungroh Yoon.
\newblock Entropy is not enough for test-time adaptation: From the perspective of disentangled factors.
\newblock \emph{arXiv preprint arXiv:2403.07366}, 2024.

\bibitem[Li et~al.(2017)Li, Yang, Song, and Hospedales]{PACS}
Da Li, Yongxin Yang, Yi-Zhe Song, and Timothy~M. Hospedales.
\newblock Deeper, broader and artier domain generalization, 2017.

\bibitem[Li et~al.(2022)Li, Li, Xiong, and Hoi]{vlm4}
Junnan Li, Dongxu Li, Caiming Xiong, and Steven Hoi.
\newblock Blip: Bootstrapping language-image pre-training for unified vision-language understanding and generation.
\newblock In \emph{International conference on machine learning}, pages 12888--12900. PMLR, 2022.

\bibitem[Li et~al.(2023{\natexlab{a}})Li, Li, Savarese, and Hoi]{vlm5}
Junnan Li, Dongxu Li, Silvio Savarese, and Steven Hoi.
\newblock Blip-2: Bootstrapping language-image pre-training with frozen image encoders and large language models.
\newblock In \emph{International conference on machine learning}, pages 19730--19742. PMLR, 2023{\natexlab{a}}.

\bibitem[Li et~al.(2024{\natexlab{a}})Li, Yang, Zhao, Wang, Wang, Yang, Sun, Kou, Qian, and Zhang]{li2024correlation}
Mingcheng Li, Dingkang Yang, Xiao Zhao, Shuaibing Wang, Yan Wang, Kun Yang, Mingyang Sun, Dongliang Kou, Ziyun Qian, and Lihua Zhang.
\newblock Correlation-decoupled knowledge distillation for multimodal sentiment analysis with incomplete modalities.
\newblock In \emph{Proceedings of the IEEE/CVF Conference on Computer Vision and Pattern Recognition}, pages 12458--12468, 2024{\natexlab{a}}.

\bibitem[Li et~al.(2023{\natexlab{b}})Li, Li, Yang, Zhao, Song, Luo, Li, and Yang]{li2023curriculum}
Zheng Li, Xiang Li, Lingfeng Yang, Borui Zhao, Renjie Song, Lei Luo, Jun Li, and Jian Yang.
\newblock Curriculum temperature for knowledge distillation.
\newblock In \emph{Proceedings of the AAAI Conference on Artificial Intelligence}, pages 1504--1512, 2023{\natexlab{b}}.

\bibitem[Li et~al.(2024{\natexlab{b}})Li, Li, Fu, Zhang, Wang, Chen, and Yang]{li2024promptkd}
Zheng Li, Xiang Li, Xinyi Fu, Xin Zhang, Weiqiang Wang, Shuo Chen, and Jian Yang.
\newblock Promptkd: Unsupervised prompt distillation for vision-language models.
\newblock In \emph{Proceedings of the IEEE/CVF Conference on Computer Vision and Pattern Recognition}, pages 26617--26626, 2024{\natexlab{b}}.

\bibitem[Lin et~al.(2022)Lin, Geng, Zhang, Gao, De~Melo, Wang, Dai, Qiao, and Li]{lin2022frozen}
Ziyi Lin, Shijie Geng, Renrui Zhang, Peng Gao, Gerard De~Melo, Xiaogang Wang, Jifeng Dai, Yu Qiao, and Hongsheng Li.
\newblock Frozen clip models are efficient video learners.
\newblock In \emph{European Conference on Computer Vision}, pages 388--404. Springer, 2022.

\bibitem[Liu et~al.(2023{\natexlab{a}})Liu, Li, Wu, and Lee]{liu2023llava}
Haotian Liu, Chunyuan Li, Qingyang Wu, and Yong~Jae Lee.
\newblock Visual instruction tuning, 2023{\natexlab{a}}.

\bibitem[Liu et~al.(2023{\natexlab{b}})Liu, Yang, Jia, Lu, Guo, Xue, and Zhang]{vida}
Jiaming Liu, Senqiao Yang, Peidong Jia, Ming Lu, Yandong Guo, Wei Xue, and Shanghang Zhang.
\newblock Vida: Homeostatic visual domain adapter for continual test time adaptation.
\newblock \emph{arXiv preprint arXiv:2306.04344}, 2023{\natexlab{b}}.

\bibitem[Lu et~al.(2019)Lu, Batra, Parikh, and Lee]{lu2019vilbert}
Jiasen Lu, Dhruv Batra, Devi Parikh, and Stefan Lee.
\newblock Vilbert: Pretraining task-agnostic visiolinguistic representations for vision-and-language tasks.
\newblock \emph{Advances in neural information processing systems}, 32, 2019.

\bibitem[Maharana et~al.(2025)Maharana, Zhang, Karlinsky, Feris, and Guo]{maharana2025batclip}
Sarthak Maharana, Baoming Zhang, Leonid Karlinsky, Rogerio Feris, and Yunhui Guo.
\newblock Batclip: Bimodal online test-time adaptation for clip.
\newblock In \emph{Proceedings of the IEEE/CVF International Conference on Computer Vision}, pages 1569--1579, 2025.

\bibitem[McCoy et~al.(2019)McCoy, Pavlick, and Linzen]{shift_2}
R~Thomas McCoy, Ellie Pavlick, and Tal Linzen.
\newblock Right for the wrong reasons: Diagnosing syntactic heuristics in natural language inference.
\newblock \emph{arXiv preprint arXiv:1902.01007}, 2019.

\bibitem[Niu et~al.(2022)Niu, Wu, Zhang, Chen, Zheng, Zhao, and Tan]{EATA}
Shuaicheng Niu, Jiaxiang Wu, Yifan Zhang, Yaofo Chen, Shijian Zheng, Peilin Zhao, and Mingkui Tan.
\newblock Efficient test-time model adaptation without forgetting.
\newblock In \emph{International conference on machine learning}, pages 16888--16905. PMLR, 2022.

\bibitem[Niu et~al.(2023)Niu, Wu, Zhang, Wen, Chen, Zhao, and Tan]{SAR}
Shuaicheng Niu, Jiaxiang Wu, Yifan Zhang, Zhiquan Wen, Yaofo Chen, Peilin Zhao, and Mingkui Tan.
\newblock Towards stable test-time adaptation in dynamic wild world.
\newblock \emph{arXiv preprint arXiv:2302.12400}, 2023.

\bibitem[Oh et~al.(2024)Oh, Li, Song, Yun, and Han]{oh2024dawin}
Changdae Oh, Yixuan Li, Kyungwoo Song, Sangdoo Yun, and Dongyoon Han.
\newblock Dawin: Training-free dynamic weight interpolation for robust adaptation.
\newblock \emph{arXiv preprint arXiv:2410.03782}, 2024.

\bibitem[Park et~al.(2024)Park, Gupta, and Wong]{park2024test}
Hyoungseob Park, Anjali Gupta, and Alex Wong.
\newblock Test-time adaptation for depth completion.
\newblock In \emph{Proceedings of the IEEE/CVF Conference on Computer Vision and Pattern Recognition}, pages 20519--20529, 2024.

\bibitem[Pei et~al.(2023)Pei, Liu, Li, Shao, Xu, Dai, Lu, and Yan]{pei2023clipping}
Renjing Pei, Jianzhuang Liu, Weimian Li, Bin Shao, Songcen Xu, Peng Dai, Juwei Lu, and Youliang Yan.
\newblock Clipping: Distilling clip-based models with a student base for video-language retrieval.
\newblock In \emph{Proceedings of the IEEE/CVF Conference on Computer Vision and Pattern Recognition}, pages 18983--18992, 2023.

\bibitem[Radford et~al.(2021)Radford, Kim, Hallacy, Ramesh, Goh, Agarwal, Sastry, Askell, Mishkin, Clark, et~al.]{clip}
Alec Radford, Jong~Wook Kim, Chris Hallacy, Aditya Ramesh, Gabriel Goh, Sandhini Agarwal, Girish Sastry, Amanda Askell, Pamela Mishkin, Jack Clark, et~al.
\newblock Learning transferable visual models from natural language supervision.
\newblock In \emph{International conference on machine learning}, pages 8748--8763. PmLR, 2021.

\bibitem[Sahoo et~al.(2025)Sahoo, ElAraby, Ngnawe, Pequignot, Precioso, and Gagn{\'e}]{domaindetector}
Sabyasachi Sahoo, Mostafa ElAraby, Jonas Ngnawe, Yann~Batiste Pequignot, Fr{\'e}d{\'e}ric Precioso, and Christian Gagn{\'e}.
\newblock A layer selection approach to test time adaptation.
\newblock In \emph{Proceedings of the AAAI Conference on Artificial Intelligence}, pages 20237--20245, 2025.

\bibitem[Schneider et~al.(2020)Schneider, Rusak, Eck, Bringmann, Brendel, and Bethge]{bn_adapt}
Steffen Schneider, Evgenia Rusak, Luisa Eck, Oliver Bringmann, Wieland Brendel, and Matthias Bethge.
\newblock Removing covariate shift improves robustness against common corruptions.
\newblock \emph{CoRR}, abs/2006.16971, 2020.

\bibitem[Shu et~al.(2022)Shu, Nie, Huang, Yu, Goldstein, Anandkumar, and Xiao]{tpt}
Manli Shu, Weili Nie, De-An Huang, Zhiding Yu, Tom Goldstein, Anima Anandkumar, and Chaowei Xiao.
\newblock Test-time prompt tuning for zero-shot generalization in vision-language models.
\newblock \emph{Advances in Neural Information Processing Systems}, 35:\penalty0 14274--14289, 2022.

\bibitem[Silva-Rodr{\'\i}guez et~al.(2025)Silva-Rodr{\'\i}guez, Ben~Ayed, and Dolz]{silva2025conformal}
Julio Silva-Rodr{\'\i}guez, Ismail Ben~Ayed, and Jose Dolz.
\newblock Conformal prediction for zero-shot models.
\newblock In \emph{Proceedings of the Computer Vision and Pattern Recognition Conference}, pages 19931--19941, 2025.

\bibitem[Simons et~al.(2023)Simons, Raychaudhuri, Ahmed, You, Karydis, and Roy-Chowdhury]{simons2023summitsourcefreeadaptationunimodal}
Cody Simons, Dripta~S. Raychaudhuri, Sk~Miraj Ahmed, Suya You, Konstantinos Karydis, and Amit~K. Roy-Chowdhury.
\newblock Summit: Source-free adaptation of uni-modal models to multi-modal targets, 2023.

\bibitem[Sze et~al.(2017)Sze, Chen, Yang, and Emer]{DNN}
Vivienne Sze, Yu-Hsin Chen, Tien-Ju Yang, and Joel~S Emer.
\newblock Efficient processing of deep neural networks: A tutorial and survey.
\newblock \emph{Proceedings of the IEEE}, 105\penalty0 (12):\penalty0 2295--2329, 2017.

\bibitem[Venkateswara et~al.(2017)Venkateswara, Eusebio, Chakraborty, and Panchanathan]{officehome}
Hemanth Venkateswara, Jose Eusebio, Shayok Chakraborty, and Sethuraman Panchanathan.
\newblock Deep hashing network for unsupervised domain adaptation.
\newblock In \emph{Proceedings of the IEEE Conference on Computer Vision and Pattern Recognition}, pages 5018--5027, 2017.

\bibitem[Villani et~al.(2008)]{ot2}
C{\'e}dric Villani et~al.
\newblock \emph{Optimal transport: old and new}.
\newblock Springer, 2008.

\bibitem[Wang et~al.(2020)Wang, Shelhamer, Liu, Olshausen, and Darrell]{Tent}
Dequan Wang, Evan Shelhamer, Shaoteng Liu, Bruno Olshausen, and Trevor Darrell.
\newblock Tent: Fully test-time adaptation by entropy minimization.
\newblock \emph{arXiv preprint arXiv:2006.10726}, 2020.

\bibitem[Wang et~al.(2022)Wang, Fink, Van~Gool, and Dai]{cotta}
Qin Wang, Olga Fink, Luc Van~Gool, and Dengxin Dai.
\newblock Continual test-time domain adaptation.
\newblock In \emph{Proceedings of the IEEE/CVF Conference on Computer Vision and Pattern Recognition}, pages 7201--7211, 2022.

\bibitem[Wu et~al.(2024)Wu, Zhang, Li, Chen, Liang, Yang, and Li]{wu2024cascade}
Ge Wu, Xin Zhang, Zheng Li, Zhaowei Chen, Jiajun Liang, Jian Yang, and Xiang Li.
\newblock Cascade prompt learning for vision-language model adaptation.
\newblock In \emph{European Conference on Computer Vision}, pages 304--321. Springer, 2024.

\bibitem[Wu et~al.(2023)Wu, Peng, Zhou, Xiao, Liu, Yuan, Xuan, Valenzuela, Chen, Wang, et~al.]{wu2023tinyclip}
Kan Wu, Houwen Peng, Zhenghong Zhou, Bin Xiao, Mengchen Liu, Lu Yuan, Hong Xuan, Michael Valenzuela, Xi~Stephen Chen, Xinggang Wang, et~al.
\newblock Tinyclip: Clip distillation via affinity mimicking and weight inheritance.
\newblock In \emph{Proceedings of the IEEE/CVF International Conference on Computer Vision}, pages 21970--21980, 2023.

\bibitem[Xiong et~al.(2024)Xiong, Yang, Song, Wang, and Xu]{xiong2024modality}
Baochen Xiong, Xiaoshan Yang, Yaguang Song, Yaowei Wang, and Changsheng Xu.
\newblock Modality-collaborative test-time adaptation for action recognition.
\newblock In \emph{Proceedings of the IEEE/CVF Conference on Computer Vision and Pattern Recognition}, pages 26732--26741, 2024.

\bibitem[Yang et~al.(2022{\natexlab{a}})Yang, An, Cai, and Xu]{yang2022mutual}
Chuanguang Yang, Zhulin An, Linhang Cai, and Yongjun Xu.
\newblock Mutual contrastive learning for visual representation learning.
\newblock In \emph{Proceedings of the AAAI Conference on Artificial Intelligence}, pages 3045--3053, 2022{\natexlab{a}}.

\bibitem[Yang et~al.(2022{\natexlab{b}})Yang, An, Zhou, Cai, Zhi, Wu, Xu, and Zhang]{yang2022mixskd}
Chuanguang Yang, Zhulin An, Helong Zhou, Linhang Cai, Xiang Zhi, Jiwen Wu, Yongjun Xu, and Qian Zhang.
\newblock Mixskd: Self-knowledge distillation from mixup for image recognition.
\newblock In \emph{European Conference on Computer Vision}, pages 534--551. Springer, 2022{\natexlab{b}}.

\bibitem[Yang et~al.(2023)Yang, An, Zhou, Zhuang, Xu, and Zhang]{yang2023online}
Chuanguang Yang, Zhulin An, Helong Zhou, Fuzhen Zhuang, Yongjun Xu, and Qian Zhang.
\newblock Online knowledge distillation via mutual contrastive learning for visual recognition.
\newblock \emph{IEEE Transactions on Pattern Analysis and Machine Intelligence}, 45\penalty0 (8):\penalty0 10212--10227, 2023.

\bibitem[Yang et~al.(2024)Yang, An, Huang, Bi, Yu, Yang, Diao, and Xu]{yang2024clip}
Chuanguang Yang, Zhulin An, Libo Huang, Junyu Bi, Xinqiang Yu, Han Yang, Boyu Diao, and Yongjun Xu.
\newblock Clip-kd: An empirical study of clip model distillation.
\newblock In \emph{Proceedings of the IEEE/CVF Conference on Computer Vision and Pattern Recognition}, pages 15952--15962, 2024.

\bibitem[Yosinski et~al.(2014)Yosinski, Clune, Bengio, and Lipson]{shift_1}
Jason Yosinski, Jeff Clune, Yoshua Bengio, and Hod Lipson.
\newblock How transferable are features in deep neural networks?
\newblock \emph{Advances in neural information processing systems}, 27, 2014.

\bibitem[Yu et~al.(2019)Yu, Yu, Cui, Tao, and Tian]{yu2019deep}
Zhou Yu, Jun Yu, Yuhao Cui, Dacheng Tao, and Qi Tian.
\newblock Deep modular co-attention networks for visual question answering.
\newblock In \emph{Proceedings of the IEEE/CVF conference on computer vision and pattern recognition}, pages 6281--6290, 2019.

\bibitem[Yuan et~al.(2023)Yuan, Xie, and Li]{rotta}
Longhui Yuan, Binhui Xie, and Shuang Li.
\newblock Robust test-time adaptation in dynamic scenarios.
\newblock In \emph{Proceedings of the IEEE/CVF Conference on Computer Vision and Pattern Recognition}, pages 15922--15932, 2023.

\bibitem[Zhai et~al.(2023)Zhai, Mustafa, Kolesnikov, and Beyer]{siglip}
Xiaohua Zhai, Basil Mustafa, Alexander Kolesnikov, and Lucas Beyer.
\newblock Sigmoid loss for language image pre-training.
\newblock In \emph{Proceedings of the IEEE/CVF international conference on computer vision}, pages 11975--11986, 2023.

\bibitem[Zhang et~al.(2022{\natexlab{a}})Zhang, Levine, and Finn]{Memo}
Marvin Zhang, Sergey Levine, and Chelsea Finn.
\newblock Memo: Test time robustness via adaptation and augmentation.
\newblock \emph{Advances in neural information processing systems}, 35:\penalty0 38629--38642, 2022{\natexlab{a}}.

\bibitem[Zhang et~al.(2022{\natexlab{b}})Zhang, Zeng, Guo, and Li]{zhang2022can}
Renrui Zhang, Ziyao Zeng, Ziyu Guo, and Yafeng Li.
\newblock Can language understand depth?
\newblock In \emph{Proceedings of the 30th ACM International Conference on Multimedia}, pages 6868--6874, 2022{\natexlab{b}}.

\bibitem[Zhang et~al.(2023)Zhang, Wang, Qiao, Gao, and Li]{zhang2023learning}
Renrui Zhang, Liuhui Wang, Yu Qiao, Peng Gao, and Hongsheng Li.
\newblock Learning 3d representations from 2d pre-trained models via image-to-point masked autoencoders.
\newblock In \emph{Proceedings of the IEEE/CVF Conference on Computer Vision and Pattern Recognition}, pages 21769--21780, 2023.

\bibitem[Zhang et~al.(2024{\natexlab{a}})Zhang, Wang, Guo, Dai, Chen, and Xia]{boostadapter}
Taolin Zhang, Jinpeng Wang, Hang Guo, Tao Dai, Bin Chen, and Shu-Tao Xia.
\newblock Boostadapter: Improving test-time adaptation via regional bootstrapping.
\newblock \emph{arXiv preprint arXiv:2410.15430}, 2024{\natexlab{a}}.

\bibitem[Zhang et~al.(2024{\natexlab{b}})Zhang, Li, Chu, Hai, Xu, Yang, Guan, Xu, and Cui]{zhang2024out}
Xingxuan Zhang, Jiansheng Li, Wenjing Chu, Junjia Hai, Renzhe Xu, Yuqing Yang, Shikai Guan, Jiazheng Xu, and Peng Cui.
\newblock On the out-of-distribution generalization of multimodal large language models.
\newblock \emph{arXiv preprint arXiv:2402.06599}, 2024{\natexlab{b}}.

\bibitem[Zhang et~al.(2024{\natexlab{c}})Zhang, Mehra, Niu, and Hamm]{dpcore}
Yunbei Zhang, Akshay Mehra, Shuaicheng Niu, and Jihun Hamm.
\newblock Dpcore: Dynamic prompt coreset for continual test-time adaptation.
\newblock \emph{arXiv preprint arXiv:2406.10737}, 2024{\natexlab{c}}.

\bibitem[Zhang et~al.(2024{\natexlab{d}})Zhang, Unell, Wang, Ghosh, Su, Schmidt, and Yeung-Levy]{zhang2024visually}
Yuhui Zhang, Alyssa Unell, Xiaohan Wang, Dhruba Ghosh, Yuchang Su, Ludwig Schmidt, and Serena Yeung-Levy.
\newblock Why are visually-grounded language models bad at image classification?
\newblock \emph{Advances in Neural Information Processing Systems}, 37:\penalty0 51727--51753, 2024{\natexlab{d}}.

\bibitem[Zhang et~al.(2022{\natexlab{c}})Zhang, Meng, Wang, Jiang, Liu, and Yang]{zhang2022unims}
Zhengkun Zhang, Xiaojun Meng, Yasheng Wang, Xin Jiang, Qun Liu, and Zhenglu Yang.
\newblock Unims: A unified framework for multimodal summarization with knowledge distillation.
\newblock In \emph{Proceedings of the AAAI conference on artificial intelligence}, pages 11757--11764, 2022{\natexlab{c}}.

\bibitem[Zhao et~al.(2023)Zhao, Liu, Alahi, and Lin]{TTAB}
Hao Zhao, Yuejiang Liu, Alexandre Alahi, and Tao Lin.
\newblock On pitfalls of test-time adaptation.
\newblock \emph{arXiv preprint arXiv:2306.03536}, 2023.

\bibitem[Zhu et~al.(2023)Zhu, Chen, Shen, Li, and Elhoseiny]{minigpt4}
Deyao Zhu, Jun Chen, Xiaoqian Shen, Xiang Li, and Mohamed Elhoseiny.
\newblock Minigpt-4: Enhancing vision-language understanding with advanced large language models.
\newblock \emph{arXiv preprint arXiv:2304.10592}, 2023.

\bibitem[Zou et~al.(2025)Zou, Feng, Wang, Huang, Huang, Haihang, Zou, and Li]{zou2025zeroshotsubjectcentricgenerationcreative}
Kaifeng Zou, Xiaoyi Feng, Peng Wang, Tao Huang, Zizhou Huang, Zhang Haihang, Yuntao Zou, and Dagang Li.
\newblock Zero-shot subject-centric generation for creative application using entropy fusion, 2025.

\end{thebibliography}
}

% WARNING: do not forget to delete the supplementary pages from your submission 
\appendix
\clearpage
\setcounter{page}{1}
\maketitlesupplementary

\section{Pseudo-Codes of CoDiRe}

\label{app:algo}

\noindent
Algorithm~\ref{alg:CoDiRe} presents the detailed pseudo-code for the CoDiRe framework, which addresses the two empirical pitfalls in our proposed Test-Time Distillation (TTD) paradigm.

\begin{algorithm}[ht]
\caption{The detailed procedures of CoDiRe}
\label{alg:CoDiRe}
\begin{algorithmic}[1]
\Require Pretrained Target Model $f(\cdot; \theta_0)$, Frozen VLM Teacher $\mathcal{F}(\cdot)$, Unlabeled Test Data $\mathcal{X}^{\text{T}}_{t}$, Reset Threshold $\gamma_0$, Anchor Update Step $s$, Reset Ratio $\alpha$
\State Initialize current parameters $\theta_t \gets \theta_0$
\State Initialize anchor parameters $\theta^{\text{anchor}} \gets \theta_0$
\State Initialize previous parameters $\theta_{t-1} \gets \theta_0$

\For{each time step $t = 1, 2, \dots$}
    \State Receive mini-batch $\{x_i\}_{i=1}^N\subseteq \mathcal{X}^{\text{T}}_{t}$
    
    \State \textcolor{blue}{\textit{// 1. Distillation: Construct Blended Teacher}}
    \State Get logits: $z_i^{\text{tar}} = f(x_i; \theta_t)$, $z_i^{\text{tea}} = \mathcal{F}(x_i)$
    \State Compute MSP-based weight via Eq.~(2): 
    \State \quad $\lambda_i = \frac{\exp(\max p_i^{\text{tea}})}{\exp(\max p_i^{\text{tea}}) + \exp(\max p_i^{\text{tar}})}$
    \State Compute blended logits: $z_i^{\text{bt}} = \lambda_i \cdot z_i^{\text{tea}} + (1 - \lambda_i) \cdot z_i^{\text{tar}}$
    \State Get blended probability: $p_i^{\text{bt}} = \sigma(z_i^{\text{bt}})$
    
    \State \textcolor{blue}{\textit{// 2. Rectification: Optimal Transport}}
    \State Compute marginal constraints $m$ via pseudo-label voting
    \State Solve OT problem via Sinkhorn algorithm to get $\mathbf{P}^{rm}$ (Eq.~4-5)
    
    \State \textcolor{blue}{\textit{// 3. Optimization}}
    \State Compute Distillation Loss: $\mathcal{L}_{\text{Distill}}$ using $p_i^{\text{bt}}$ (Eq.~3)
    \State Compute Rectification Loss: $\mathcal{L}_{\text{Rect}}$ using $P^{rm}$ (Eq.~6)
    \State Compute Entropy Loss: $\mathcal{L}_{\text{Ent}}$ (Eq.~7)
    \State Total Loss: $\mathcal{L}_{\text{total}} = \mathcal{L}_{\text{Ent}} + \mathcal{L}_{\text{Distill}} + \mathcal{L}_{\text{Rect}}$
    \State Update parameters: $\theta_{t+1} \gets \theta_t - \eta \nabla_{\theta_t} \mathcal{L}_{\text{total}}$
    
    \State \textcolor{blue}{\textit{// 4. Distribution-Aware Reset Mechanism}}
    \State Compute displacements: $\delta_t = \theta_{t+1} - \theta_{t}$, $\delta_t^{\text{anchor}} = \theta_{t} - \theta^{\text{anchor}}$
    \State Calculate cosine similarity: $\gamma = \cos(\delta_t, \delta_t^{\text{anchor}})$
    
    \If{$\gamma < \gamma_0$} \Comment{Domain shift detected}
        \State Reset the last $\alpha\%$ layers of $\theta_{t+1}$ to $\theta_0$
    \EndIf
    
    \If{$t \bmod s == 0$} \Comment{Periodic anchor update}
        \State Update anchor: $\theta^{\text{anchor}} \gets \theta_{t+1}$
    \EndIf
    
    \State $\theta_t \gets \theta_{t+1}$
    \State $\theta_{t-1} \gets \theta_t$
\EndFor
\end{algorithmic}
\end{algorithm}

\begin{table*}[ht]
\caption{\textbf{Comparison results under corruption scenarios on CIFAR-100-C with ResNet50 as backbone.} Classification accuracy of the standard CIFAR-100 \(\rightarrow\) CIFAR-100-C online continual test-time adaptation task while continually adapting to different corruptions at the highest severity 5. The best and second-best results are highlighted in \textbf{bold} and \underline{underlined}, respectively.}
\centering
\resizebox{\textwidth}{!}{%
\begin{tabular}{c|c|ccccccccccccccc|c}
\toprule
\textbf{CIFAR-100-C} & 
\multirow{2}{*}{\textbf{Venue}} &
\multicolumn{3}{c}{\textbf{Noise}} &
\multicolumn{4}{c}{\textbf{Blur}} &
\multicolumn{4}{c}{\textbf{Weather}} &
\multicolumn{4}{c|}{\textbf{Digital}} &
\multirow{2}{*}{\textbf{Avg.}} \\ %\cline{2-16}
\textbf{ResNet50-BN} & & \textit{gauss.} & \textit{shot} & \textit{impul.} & \textit{defoc.} & \textit{glass} & \textit{motion} & \textit{zoom} & \textit{snow} & \textit{frost} & \textit{fog} & \textit{brit.} & \textit{contr.} & \textit{elastic} & \textit{pixel} & \textit{jpeg} &  \\ 
\rowcolor{gray!10} & \multicolumn{15}{c|}{\textit{Time}\hspace{2em}\textemdash\textemdash\textemdash\textemdash\textemdash\textemdash\textemdash\textemdash\textemdash\textemdash\textemdash\textemdash\textemdash\textemdash\textemdash\textemdash\textemdash\textemdash\textemdash\textemdash\textemdash\textemdash\textemdash\textemdash\textemdash\textemdash\textemdash\textemdash\textemdash\textemdash\textemdash\textemdash\textemdash\textemdash\textemdash\textemdash\textemdash\textemdash\textemdash\textemdash\textemdash\textemdash>} & \\ 
\midrule
\textbf{Source}
    & - & 10.06 & 11.67 & 6.67 & 31.87 & 17.52 & 37.58 & 36.61 & 39.01 & 27.27 & 29.71 & 60.98 & 16.87 & 41.92 & 19.26 & 41.79 & 28.59  \\

\textbf{BN Adapt} 
    & NIPS'20 & 32.13 & 32.92 & 28.77 & 59.37 & 34.97 & 56.05 & 59.35 & 46.87 & 46.29 & 49.44 & 62.46 & 58.11 & 46.09 & 47.99 & 40.60 & 46.76\(_{\pm0.01}\) \\

\textbf{Tent} 
    & ICLR'21 & 33.39 & 36.81 & 33.45 & \underline{60.64} & 39.64 & \underline{58.25} & \underline{61.97} & 51.05 & 51.44 & \underline{52.87} & 64.66 & 57.47 & 51.40 & 55.19 & 48.78 & 50.47\(_{\pm0.06}\) \\

\textbf{MEMO} 
    & NIPS'22 & 14.71 & 17.19 & 14.01 & 42.71 & 23.79 & 46.89 & 47.37 & 46.68 & 38.83 & 36.77 & 64.67 & 27.30 & 45.29 & 27.96 & 43.59 & 35.85\(_{\pm0.03}\) \\

\textbf{EATA} 
    & ICML'22 & 34.62 & 40.29 & 36.22 & 58.70 & 41.26 & 56.63 & 60.51 & 51.09 & 51.97 & 51.99 & 63.10 & 56.48 & \underline{51.44} & \underline{55.75} & \underline{50.49} & 50.70\(_{\pm0.07}\) \\

\textbf{SAR} 
    & ICLR'23 & 33.64 & 37.25 & 33.33 & 59.05 & 38.27 & 56.15 & 59.34 & 48.98 & 49.04 & 50.65 & 62.21 & 55.52 & 49.28 & 52.59 & 46.67 & 48.80\(_{\pm0.15}\) \\

\textbf{DeYO} 
    & ICLR'24 & \underline{37.07} & \underline{45.81} & 39.81 & 58.39 & \underline{42.80} & 56.43 & 60.03 & 50.95 & 51.82 & 51.97 & 61.09 & 55.36 & 50.82 & 54.86 & 49.51 & \underline{51.11}\(_{\pm0.51}\) \\

\midrule
\textbf{CoTTA} 
    & CVPR'22 & 31.45 & 31.20 & 27.65 & 58.68 & 33.53 & 55.36 & 58.36 & 45.61 & 44.69 & 46.45 & 61.70 & 57.88 & 44.33 & 44.89 & 38.62 & 45.36\(_{\pm0.13}\)  \\

\textbf{NOTE} 
    & NIPS'22 & 23.57 & 35.33 & 25.67 & 16.40 & 28.11 & 38.28 & 53.01 & 36.64 & 47.34 & 36.72 & 57.57 & 38.90 & 34.10 & 25.18 & 34.90 & 35.45\(_{\pm0.14}\)  \\

\textbf{RoTTA} 
    & CVPR'23 & 27.09 & 26.39 & 25.46 & 49.02 & 27.59 & 47.82 & 52.86 & 39.05 & 33.46 & 40.71 & 56.34 & 36.37 & 41.80 & 40.11 & 38.20 & 38.82\(_{\pm0.19}\)  \\
    
\textbf{SANTA} 
    & TMLR'23 & 33.60 & 37.42 & 33.60 & 60.39 & 39.17 & 57.04 & 60.43 & 50.29 & 50.40 & 51.61 & 64.27 & 56.87 & 49.85 & 53.12 & 47.61 & 49.71\(_{\pm0.33}\)  \\
    
\textbf{ViDA} 
    & ICLR'24 & 32.25 & 33.38 & 29.70 & 59.64 & 36.06 & 57.14 & 60.28 & 47.74 & 47.49 & 50.84 & 63.30 & \underline{58.93} & 48.18 & 50.44 & 43.98 & 47.96\(_{\pm0.09}\)  \\

\midrule

\textbf{CLIP} 
    & ICML'21 & 36.14 & 38.31 & \underline{43.25} & 43.00 & 24.27 & 45.95 & 49.14 & \underline{54.44} & \underline{57.12} & 39.57 & \underline{65.74} & 38.00 & 37.49 & 46.45 & 42.26 & 44.08 \\

\rowcolor{cyan!10}\textbf{Ours} 
    & - & \textbf{51.70} & \textbf{58.43} & \textbf{57.45} & \textbf{66.69} & \textbf{47.08} & \textbf{66.37} & \textbf{70.56} & \textbf{66.22} & \textbf{68.37} & \textbf{61.79} & \textbf{76.91} & \textbf{64.91} & \textbf{57.93} & \textbf{65.11} & \textbf{59.68} & \textbf{62.61}\(_{\pm0.13}\)  \\

\bottomrule
\end{tabular}%
}
\label{tab:cifar100c}
\end{table*}
\section{MSP-Accuracy Bninning Experiment}
\label{app:msp_exp}
This section presents an empirical validation of the reliability of Maximum Softmax Probability (MSP) as a confidence metric under distribution shifts. This experiment serves as the empirical foundation for the theoretical analysis presented in Appendix~\ref{app:theory}, specifically supporting the error bound assumption in Eq.~(\ref{eq:error_bound}).

\paragraph{Experimental Setup.}
To rigorously assess the calibration of MSP, we conduct evaluations on two benchmark datasets: CIFAR-10-C and ImageNet-C. 
We pretrain ResNet-50 and ViT-B/16 as target models on CIFAR-10 and ImageNet, respectively, alongside the CLIP-ViT-L/14 teacher model.
The evaluation encompasses all 15 corruption types at the highest severity level 5.
For each model, we aggregate the prediction results across all corruption scenarios. 
We then discretize the MSP-based confidence scores into bins with an interval of 0.05. 
For each bin, we calculate the average accuracy of the samples falling within that confidence range.

\begin{figure}[htbp]
    \centering
    \begin{subfigure}{0.49\columnwidth}
        \centering
        \includegraphics[width=\textwidth]{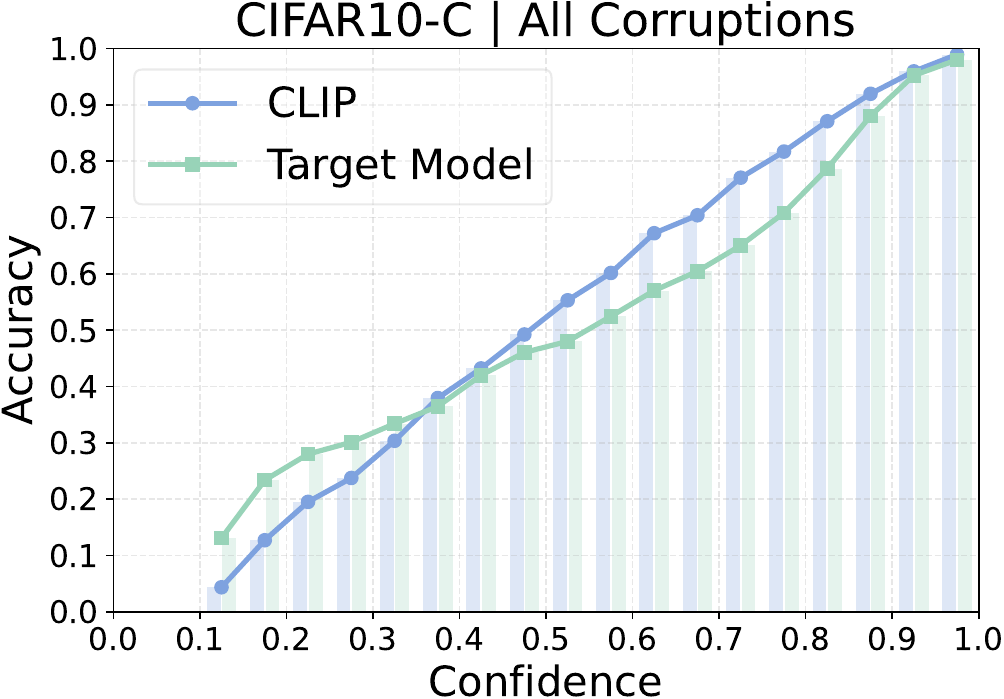}
        \label{fig:confbin_cifar10}
    \end{subfigure}
    \begin{subfigure}{0.49\columnwidth}
        \centering
        \includegraphics[width=\textwidth]{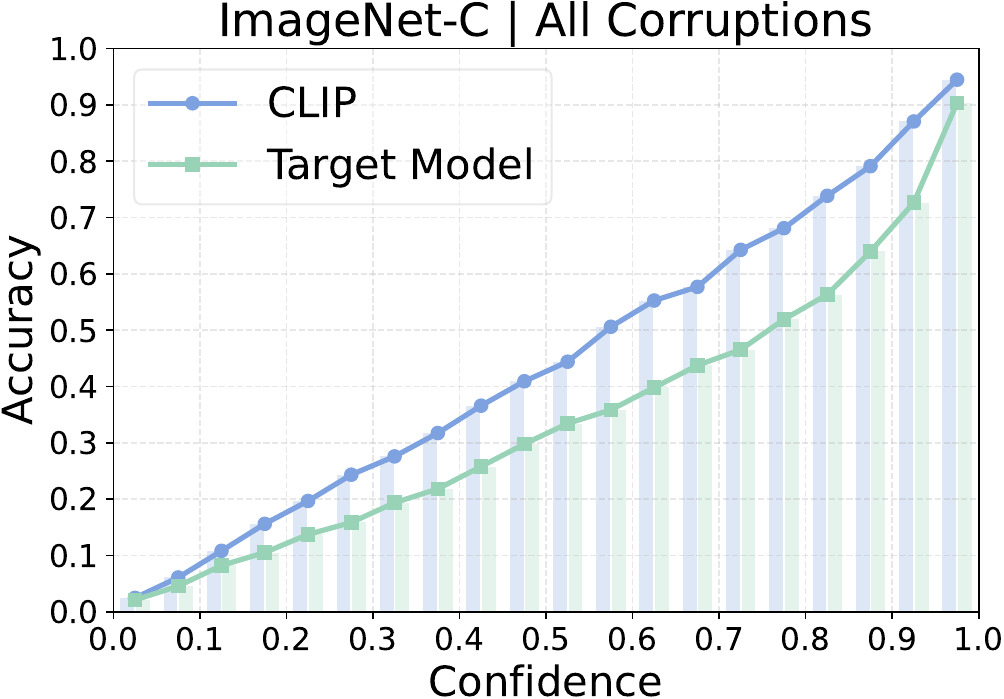}
        \label{fig:confbin_imagenet}
    \end{subfigure}
    \vspace{-18pt}
    \caption{MSP-Accuracy Binning Plots on CIFAR-10-C and ImageNet-C. The plots aggregate results across all 15 corruption types. The x-axis represents the MSP-based confidence bins, and the y-axis represents the average accuracy within each bin. The strong alignment in the high-confidence region validates the assumption that high MSP implies high accuracy.}
    \label{fig:conf_bin}
    \vspace{-18pt}
\end{figure}

\paragraph{Results and Analysis.}
Figure~\ref{fig:conf_bin} illustrates the relationship between confidence and accuracy for both the target model and CLIP.
We observe a consistent, monotonic positive correlation between MSP and classification accuracy across both datasets and models. 
This indicates that MSP serves as an effective proxy for the model's correctness likelihood, even under severe distribution shifts.

Crucially, the results demonstrate that in the high-confidence regime, the models exhibit extremely high reliability. 
Specifically, as the confidence score approaches 1.0, the accuracy asymptotically approaches 100\%. 
For instance, on CIFAR-10-C, samples with confidence scores above 0.95 achieve near-perfect accuracy for both models. 
This empirical observation strongly supports our theoretical premise: there exists a confidence threshold $\mu$ above which the probability of an incorrect prediction is negligible (bounded by a small $\alpha$). 
This also justifies our strategy of utilizing high-MSP predictions to construct a reliable blended teacher signal.

\begin{table*}[ht]
\caption{\textbf{Comparison results under corruption scenarios on CIFAR-10-C with ResNet26 as backbone.} Classification accuracy of the standard CIFAR-10 \(\rightarrow\) CIFAR-10-C online continual test-time adaptation task while continually adapting to different corruptions at the highest severity 5. The best and second-best results are highlighted in \textbf{bold} and \underline{underlined}, respectively.}
\centering
\resizebox{\textwidth}{!}{%
\begin{tabular}{c|c|ccccccccccccccc|c}
\toprule
\textbf{CIFAR-10-C} &
\multirow{2}{*}{\textbf{Venue}} &
\multicolumn{3}{c}{\textbf{Noise}} &
\multicolumn{4}{c}{\textbf{Blur}} &
\multicolumn{4}{c}{\textbf{Weather}} &
\multicolumn{4}{c|}{\textbf{Digital}} &
\multirow{2}{*}{\textbf{Avg.}} \\ %\cline{2-16}
\textbf{ResNet26-BN} & & \textit{gauss.} & \textit{shot} & \textit{impul.} & \textit{defoc.} & \textit{glass} & \textit{motion} & \textit{zoom} & \textit{snow} & \textit{frost} & \textit{fog} & \textit{brit.} & \textit{contr.} & \textit{elastic} & \textit{pixel} & \textit{jpeg} &  \\ 
\rowcolor{gray!10} & \multicolumn{15}{c|}{\textit{Time}\hspace{2em}\textemdash\textemdash\textemdash\textemdash\textemdash\textemdash\textemdash\textemdash\textemdash\textemdash\textemdash\textemdash\textemdash\textemdash\textemdash\textemdash\textemdash\textemdash\textemdash\textemdash\textemdash\textemdash\textemdash\textemdash\textemdash\textemdash\textemdash\textemdash\textemdash\textemdash\textemdash\textemdash\textemdash\textemdash\textemdash\textemdash\textemdash\textemdash\textemdash\textemdash\textemdash\textemdash>} & \\ 
\midrule
\textbf{Source} 
    & - & 26.89 & 33.22 & 30.54 & 59.83 & 48.12 & 62.12 & 61.96 & 74.51 & 60.68 & 60.55 & 89.90 & 44.77 & 73.25 & 38.57 & 69.08 & 55.60  \\

\textbf{BN Adapt} 
    & NIPS'20 & 60.70 & 63.02 & 53.90 & 82.69 & 58.89 & 80.18 & 82.25 & 74.83 & 74.57 & 79.31 & 85.97 & 82.28 & 71.08 & 73.42 & 64.53 & 72.51\(_{\pm0.05}\)  \\

\textbf{Tent}
    & ICLR'21 & 62.04 & 66.34 & 58.61 & 83.82 & 62.88 & 81.50 & 83.80 & 77.81 & 78.05 & 80.41 & 87.06 & 82.23 & 75.01 & 78.65 & 71.21 & 75.29\(_{\pm0.07}\)  \\

\textbf{MEMO}
    & NIPS'22 & 39.15 & 46.58 & 49.22 & 70.52 & 56.15 & 71.22 & 70.98 & 79.91 & 72.12 & 70.09 & 91.21 & 62.67 & 76.35 & 45.51 & 72.59 & 64.95\(_{\pm0.02}\)  \\

\textbf{EATA}
    & ICML'22 & 60.70 & 63.02 & 53.90 & 82.69 & 58.88 & 80.18 & 82.26 & 74.84 & 74.56 & 79.31 & 85.97 & 82.28 & 71.07 & 73.43 & 64.54 & 72.51\(_{\pm0.05}\)  \\

\textbf{SAR}
    & ICLR'23 & 61.84 & 66.34 & 58.28 & 84.01 & 62.92 & 81.91 & 84.17 & 78.18 & 78.66 & 81.12 & 87.60 & 82.93 & 75.86 & 79.18 & 71.96 & 75.66\(_{\pm0.07}\)  \\

\textbf{DeYO}
    & ICLR'24 & 64.18 & \underline{71.82} & 64.57 & \underline{85.01} & \underline{67.37} & \underline{83.28} & \underline{86.03} & 81.50 & 82.75 & \underline{83.51} & 89.87 & \underline{86.96} & \underline{78.27} & \underline{83.55} & \underline{75.69} & \underline{78.96}\(_{\pm0.22}\)  \\

\midrule
\textbf{CoTTA} 
    & CVPR'22 & 60.70 & 63.02 & 53.90 & 82.69 & 58.89 & 80.18 & 82.25 & 74.83 & 74.57 & 79.31 & 85.97 & 82.28 & 71.07 & 73.42 & 64.53 & 72.51\(_{\pm0.05}\)  \\

\textbf{NOTE} 
    & NIPS'22 & 50.44 & 66.76 & 52.72 & 32.81 & 52.71 & 62.39 & 75.44 & 73.77 & 77.75 & 61.26 & 87.10 & 69.80 & 61.14 & 48.26 & 61.49 & 62.26\(_{\pm0.09}\)  \\

\textbf{RoTTA} 
    & CVPR'23 & 56.20 & 59.40 & 51.21 & 76.48 & 56.55 & 76.48 & 80.28 & 70.90 & 62.35 & 71.13 & 80.82 & 50.34 & 64.02 & 62.88 & 59.24 & 65.22\(_{\pm0.25}\)  \\

\textbf{SANTA}
    & TMLR'23 & 63.84 & 69.78 & 62.34 & 83.91 & 64.95 & 80.98 & 83.26 & 79.02 & 79.20 & 80.39 & 87.62 & 80.93 & 75.60 & 79.67 & 74.79 & 76.42\(_{\pm0.34}\)  \\

\textbf{ViDA}
    & ICLR'24 & 60.73 & 63.19 & 54.26 & 82.92 & 59.36 & 80.47 & 82.58 & 75.37 & 75.27 & 79.78 & 86.33 & 82.39 & 72.28 & 74.55 & 66.01 & 73.03\(_{\pm0.05}\)  \\

\midrule
\textbf{CLIP}
  & ICML'21 & \underline{65.39} & 66.64 & \underline{76.02} & 75.75 & 48.18 & 78.16 & 79.08
  & \underline{81.92} & \underline{84.68} & 76.14 & \underline{90.40} & 80.47 & 64.74 & 76.93 & 71.32 & 74.39 \\

\rowcolor{cyan!10}\textbf{Ours} 
    & - & \textbf{76.21} & \textbf{82.83} & \textbf{81.18} & \textbf{88.67} & \textbf{74.43} & \textbf{88.07} & \textbf{90.43} & \textbf{88.97} & \textbf{90.68} & \textbf{88.89} & \textbf{94.57} & \textbf{91.70} & \textbf{83.43} & \textbf{89.02} & \textbf{83.16} & \textbf{86.15}\(_{\pm0.01}\)  \\

\bottomrule
\end{tabular}%
}
\label{tab:cifar10c_rn26}
\end{table*}

\section{Theoretical Analysis}
\label{app:theory}

Building upon the empirical observations in Appendix~\ref{app:msp_exp}, we provide a theoretical justification for the robustness of our proposed CoDiRe framework. Specifically, we analyze why the MSP-based dynamic weighting mechanism effectively filters out noise and prioritizes accurate predictions (experts) over incorrect ones (non-experts).

\subsection{Premise and Problem Setup}

Consider a set of $M$ models $\{f_j\}_{j=1}^M$ operating on an input $x$ with ground-truth label $y$. In the context of CoDiRe, this set typically comprises the target model and the VLM teacher (i.e., $M=2$). Let $\text{Conf}_j(x) \in [0, 1]$ denote the confidence (MSP) of the $j$-th model. We define the set of \textit{expert models} for a given sample $x$ as $\mathcal{J}(x) = \{j \mid \arg\max f_j(x) = y\}$, and the set of \textit{non-expert models} as $\mathcal{J}^c(x)$.

Our analysis rests on the empirical finding that high confidence is a reliable indicator of correctness. As demonstrated in Figure~\ref{fig:conf_bin}, when the confidence exceeds a certain threshold $\mu$, the error rate is bounded by a negligible margin $\alpha$. Formally:

\begin{assumption}[Empirical Error Bound]
\label{assump:error_bound}
There exists a global confidence threshold $\mu \in (0, 1)$ and a small error bound $\alpha \approx 0$, such that for any model prediction with confidence $\text{Conf} \ge \mu$, the probability of error is bounded by:
\begin{equation}
    P(\hat{Y} \ne Y \mid \text{Conf} \ge \mu) \le \alpha.
    \label{eq:error_bound}
\end{equation}
\end{assumption}

\subsection{Dynamic Weighting Analysis}

The core mechanism of CoDiRe utilizes a soft weighting scheme (Eq.~2 in the main text) to fuse model outputs. We first establish that this mechanism inherently favors models with higher confidence.

\begin{lemma}[Confidence-Driven Expert Dominance]
\label{lemma:dominance}
Assume that for a given input $x$, expert models exhibit higher confidence than non-expert models (Confidence Dominance), i.e., $\text{Conf}_j(x) \ge \text{Conf}_k(x)$ for all $j \in \mathcal{J}(x)$ and $k \in \mathcal{J}^c(x)$. Then, the weighting mechanism assigns strictly higher importance to the experts:
\begin{equation}
    \lambda_j(x) \ge \lambda_k(x).
\end{equation}
\end{lemma}

\begin{proof}
Recall the weighting function $\lambda_i(x) = \frac{\exp(\text{Conf}_i(x))}{Z}$, where $Z$ is the normalization constant. Since the exponential function $g(z) = e^z$ is strictly monotonically increasing, the condition $\text{Conf}_j(x) \ge \text{Conf}_k(x)$ directly implies $\exp(\text{Conf}_j(x)) \ge \exp(\text{Conf}_k(x))$. Dividing by the positive constant $Z$ preserves the inequality, yielding $\lambda_j(x) \ge \lambda_k(x)$.
\end{proof}

Lemma~\ref{lemma:dominance} confirms that our blending strategy effectively aligns the teacher signal with the more confident model. However, this relies on the \textit{Confidence Dominance} assumption. Is it reasonable to assume experts are more confident? We answer this via a probabilistic analysis derived from Assumption~\ref{assump:error_bound}.

Let $E$ denote the event that a model is an expert ($\hat{Y}=Y$), $N$ denote it is a non-expert, and $H$ denote the event of high confidence ($\text{Conf} \ge \mu$). From Assumption~\ref{assump:error_bound}, we have $P(N \mid H) \le \alpha$, which implies $P(E \mid H) \ge 1 - \alpha$.

\begin{lemma}[Probabilistic Justification]
\label{lemma:prob_justification}
The likelihood of an expert model producing a high-confidence prediction is significantly higher than that of a non-expert model. Specifically, the likelihood ratio satisfies:
\begin{equation}
    \frac{P(H \mid E)}{P(H \mid N)} \ge \frac{(1 - \alpha) \cdot P(N)}{\alpha \cdot P(E)},
\end{equation}
where $P(E)$ and $P(N)$ are the global accuracy and error rates of the model, respectively.
\end{lemma}

\begin{proof}
Applying Bayes' Theorem to expand the ratio:
\begin{equation}
    \frac{P(H \mid E)}{P(H \mid N)} = \frac{P(E \mid H)P(H) / P(E)}{P(N \mid H)P(H) / P(N)} = \frac{P(E \mid H) \cdot P(N)}{P(N \mid H) \cdot P(E)}.
\end{equation}
Substituting the bounds derived from Assumption~\ref{assump:error_bound} ($P(E \mid H) \ge 1 - \alpha$ and $P(N \mid H) \le \alpha$):
\begin{equation}
    \frac{P(H \mid E)}{P(H \mid N)} \ge \frac{(1 - \alpha) \cdot P(N)}{\alpha \cdot P(E)}.
\end{equation}
\end{proof}

\noindent \textbf{Remark.} Given that $\alpha \approx 0$ in our experiments (as seen in the high-confidence bins of Figure~\ref{fig:conf_bin}), the term $\frac{1-\alpha}{\alpha}$ becomes extremely large. This indicates that $P(H \mid E) \gg P(H \mid N)$. In other words, statistically, high confidence is overwhelmingly generated by expert models. This provides a strong theoretical foundation for the Confidence Dominance assumption used in Lemma~\ref{lemma:dominance}, validating the robustness of CoDiRe's blended teacher construction.

\section{Experimental Details of Datasets, Baselines, and Hyperparameters}
\label{app:detail}

\subsection{Datasets}
\label{app:base-data}

In this paper, we evaluate the proposed method on widely recognized benchmarks that are designed to assess model robustness under distribution shifts. These benchmarks include CIFAR-10-C, ImageNet-C, Office-Home, and PACS. 
Moreover, we provide experimental results on the CIFAR-100-C dataset in Appendix \ref{app:cifar100c}. 
They encompass two types of adaptation scenarios: corruptions and domain generalizations. 
Below, we provide an overview of each dataset.

\begin{figure*}[htbp]
    \centering
    \begin{subfigure}[t]{0.19\textwidth}
        \centering
        \includegraphics[width=\textwidth]{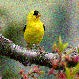}
        \caption{Gaussian Noise}
    \end{subfigure}
    \begin{subfigure}[t]{0.19\textwidth}
        \centering
        \includegraphics[width=\textwidth]{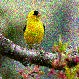}
        \caption{Shot Noise}
    \end{subfigure}
    \begin{subfigure}[t]{0.19\textwidth}
        \centering
        \includegraphics[width=\textwidth]{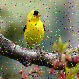}
        \caption{Impulse Noise}
    \end{subfigure}
    \begin{subfigure}[t]{0.19\textwidth}
        \centering
        \includegraphics[width=\textwidth]{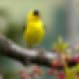}
        \caption{Defocus Blur}
    \end{subfigure}
    \begin{subfigure}[t]{0.19\textwidth}
        \centering
        \includegraphics[width=\textwidth]{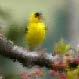}
        \caption{Glass Blur}
    \end{subfigure}
    
    \begin{subfigure}[t]{0.19\textwidth}
        \centering
        \includegraphics[width=\textwidth]{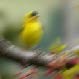}
        \caption{Motion Blur}
    \end{subfigure}
    \begin{subfigure}[t]{0.19\textwidth}
        \centering
        \includegraphics[width=\textwidth]{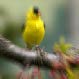}
        \caption{Zoom Blur}
    \end{subfigure}
    \begin{subfigure}[t]{0.19\textwidth}
        \centering
        \includegraphics[width=\textwidth]{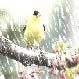}
        \caption{Snow}
    \end{subfigure}
    \begin{subfigure}[t]{0.19\textwidth}
        \centering
        \includegraphics[width=\textwidth]{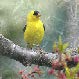}
        \caption{Frost}
    \end{subfigure}
    \begin{subfigure}[t]{0.19\textwidth}
        \centering
        \includegraphics[width=\textwidth]{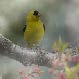}
        \caption{Fog}
    \end{subfigure}
    
    \begin{subfigure}[t]{0.19\textwidth}
        \centering
        \includegraphics[width=\textwidth]{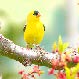}
        \caption{Brightness}
    \end{subfigure}
    \begin{subfigure}[t]{0.19\textwidth}
        \centering
        \includegraphics[width=\textwidth]{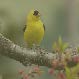}
        \caption{Contrast}
    \end{subfigure}
    \begin{subfigure}[t]{0.19\textwidth}
        \centering
        \includegraphics[width=\textwidth]{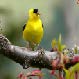}
        \caption{Elastic}
    \end{subfigure}
    \begin{subfigure}[t]{0.19\textwidth}
        \centering
        \includegraphics[width=\textwidth]{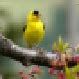}
        \caption{Pixelate}
    \end{subfigure}
    \begin{subfigure}[t]{0.19\textwidth}
        \centering
        \includegraphics[width=\textwidth]{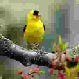}
        \caption{JPEG}
    \end{subfigure}
     \caption{\textbf{Illustration of ImageNet-C under 5 level of severity.} The dataset showcases 15 types of algorithmically generated corruptions across four categories: noise, blur, weather, and digital. Each corruption type is illustrated at five increasing levels of severity, demonstrating the progressive impact of these corruptions.}
    \label{fig:imagenetc}
\end{figure*}

\paragraph{CIFAR-10-C, CIFAR-100-C, and ImageNet-C.}
CIFAR-10-C, CIFAR-100-C, and ImageNet-C are derived from their respective base datasets, CIFAR-10, CIFAR-100, and ImageNet, by applying systematic corruptions. These datasets feature 15 corruption types, including Gaussian noise, shot noise, impulse noise, defocus blur, motion blur, zoom blur, frost, fog, brightness, contrast, elastic transformations, pixelation, and JPEG compression. 
Each corruption type is categorized into five severity levels, representing progressively more challenging distributional shifts. These datasets are widely used to benchmark model robustness against common corruptions and noise. An illustration of these corruptions is shown in Figure~\ref{fig:imagenetc}.
We consistently choose the highest level 5 in this paper.

\paragraph{Office-Home.}
Office-Home is a domain adaptation dataset consisting of images from four domains: Art, Clipart, Product, and Real-World. It contains 65 object categories commonly encountered in office and home environments, such as ``desk,'' ``keyboard,'' and ``backpack.'' The dataset is designed to evaluate domain adaptation methods under significant domain gaps, with a focus on generalizing from one domain to unseen domains. 
In this paper, we utilize "Art" as the source domain, with the other three domains as target domains. 

\paragraph{PACS.}
PACS is a domain generalization benchmark that includes images from four distinct domains: Paintings, Artistic images, Cartoons, and Sketches. The dataset spans seven object categories shared across all domains, including ``dog,'' ``guitar,'' and ``person.'' Each domain exhibits unique visual characteristics, leading to significant domain shifts. The primary evaluation task involves training on three domains and testing on the held-out domain, providing a rigorous assessment of a model's ability to generalize to unseen distributions. 
In this paper, we utilize "Art" as the source domain, with the other three domains as target domains.
\begin{table*}[ht]
\caption{\textbf{Comparison results under corruption scenarios on CIFAR-10-C with ViT-S/16 as backbone.} Classification accuracy of the standard CIFAR-10 \(\rightarrow\) CIFAR-10-C online continual test-time adaptation task while continually adapting to different corruptions at the highest severity 5. The best and second-best results are highlighted in \textbf{bold} and \underline{underlined}, respectively.}
\centering
\resizebox{\textwidth}{!}{%
\begin{tabular}{c|c|ccccccccccccccc|c}
\toprule
\textbf{CIFAR-10-C} &
\multirow{2}{*}{\textbf{Venue}} &
\multicolumn{3}{c}{\textbf{Noise}} &
\multicolumn{4}{c}{\textbf{Blur}} &
\multicolumn{4}{c}{\textbf{Weather}} &
\multicolumn{4}{c|}{\textbf{Digital}} &
\multirow{2}{*}{\textbf{Avg.}} \\ %\cline{2-16}
\textbf{VitSmall-LN} & & \textit{gauss.} & \textit{shot} & \textit{impul.} & \textit{defoc.} & \textit{glass} & \textit{motion} & \textit{zoom} & \textit{snow} & \textit{frost} & \textit{fog} & \textit{brit.} & \textit{contr.} & \textit{elastic} & \textit{pixel} & \textit{jpeg} &  \\ 
\rowcolor{gray!10} & \multicolumn{15}{c|}{\textit{Time}\hspace{2em}\textemdash\textemdash\textemdash\textemdash\textemdash\textemdash\textemdash\textemdash\textemdash\textemdash\textemdash\textemdash\textemdash\textemdash\textemdash\textemdash\textemdash\textemdash\textemdash\textemdash\textemdash\textemdash\textemdash\textemdash\textemdash\textemdash\textemdash\textemdash\textemdash\textemdash\textemdash\textemdash\textemdash\textemdash\textemdash\textemdash\textemdash\textemdash\textemdash\textemdash\textemdash\textemdash>} & \\ 
\midrule
\textbf{Source} 
    & - & 66.53 & 71.67 & 82.21 & 94.18 & 77.43 & 89.42 & 95.08 & 94.50 & 92.65 & 87.15 & 97.11 & 89.97 & 87.27 & 75.56 & 84.51 & 85.68  \\

\textbf{Tent}
    & ICLR'21 & 69.98 & 77.88 & 84.47 & \underline{94.57} & 78.50 & 90.75 & 95.87 & 94.72 & 93.22 & 89.45 & 97.08 & 92.39 & 88.66 & 69.89 & 85.32 & 86.85\(_{\pm0.03}\)  \\

\textbf{MEMO}
    & NIPS'22 & 69.43 & 74.23 & 82.74 & 94.46 & 78.36 & 89.92 & 95.27 & 94.71 & 93.25 & 88.68 & \underline{97.39} & 91.29 & 88.14 & 79.26 & 85.78 & 86.86\(_{\pm0.02}\)  \\

\textbf{EATA}
    & ICML'22 & 66.54 & 71.71 & 82.22 & 94.18 & 77.43 & 89.42 & 95.09 & 94.50 & 92.66 & 87.16 & 97.11 & 89.97 & 87.27 & 75.55 & 84.53 & 85.69\(_{\pm0.00}\)  \\

\textbf{SAR}
    & ICLR'23 & 70.06 & 78.20 & 84.61 & {94.63} & 78.42 & 91.01 & 95.88 & 94.80 & 93.45 & 89.44 & 97.07 & 92.29 & 88.76 & 67.49 & 85.12 & 86.75\(_{\pm0.05}\)  \\

\textbf{DeYO}
    & ICLR'24 & 74.77 & 81.81 & 86.89 & \textbf{94.68} & 77.87 & \underline{92.12} & \underline{95.95} & \underline{95.04} & 94.11 & \underline{92.48} & 96.73 & \underline{95.61} & 87.22 & 85.67 & 83.78 & 88.98\(_{\pm0.03}\)  \\

\midrule
\textbf{CoTTA} 
    & CVPR'22 & 66.53 & 71.67 & 82.21 & 94.18 & 77.43 & 89.42 & 95.08 & 94.50 & 92.65 & 87.15 & 97.11 & 89.97 & 87.26 & 75.56 & 84.51 & 85.68\(_{\pm0.00}\)  \\

\textbf{NOTE} 
    & NIPS'22 & 67.36 & 73.74 & 83.03 & 94.29 & 78.11 & 89.97 & 95.39 & 94.67 & 93.02 & 87.51 & 97.25 & 90.57 & 87.92 & 73.60 & 85.25 & 86.11\(_{\pm0.00}\)  \\

\textbf{RoTTA} 
    & CVPR'23 & 66.62 & 71.64 & 83.32 & 94.34 & 76.20 & 90.14 & 95.11 & 94.03 & 90.94 & 83.74 & 96.46 & 89.08 & 85.78 & 72.30 & 71.71 & 84.09\(_{\pm0.12}\)  \\

\textbf{SANTA}
    & TMLR'23 & 66.03 & 70.73 & 83.47 & 94.29 & 77.63 & 89.86 & 95.65 & 94.67 & 92.29 & 86.79 & 96.84 & 89.77 & 88.78 & 75.34 & 83.17 & 85.69\(_{\pm0.06}\)  \\

\textbf{ViDA}
    & ICLR'24 & 66.57 & 71.50 & 82.53 & 94.24 & 77.34 & 89.60 & 95.17 & 94.51 & 92.76 & 87.40 & 97.23 & 90.56 & 87.58 & 75.97 & 84.69 & 85.84\(_{\pm0.06}\)  \\

\textbf{DPCore} 
    & ICLR'25 & \underline{81.07} & \underline{84.61} & \textbf{88.52} & 94.54 & \underline{82.01} & 91.76 & 95.65 & 94.79 & \underline{94.37} & 91.59 & 96.83 & 94.84 & \underline{89.17} & \underline{92.21} & \underline{86.01} & \underline{90.53}\(_{\pm0.04}\)  \\

\midrule
\textbf{CLIP}
  & ICML'21 & 65.39 & 66.64 & 76.02 & 75.75 & 48.18 & 78.16 & 79.08
  & 81.92 & 84.68 & 76.14 & 90.40 & 80.47 & 64.74 & 76.93 & 71.32 & 74.39 \\

\rowcolor{cyan!10}\textbf{Ours} 
    & - & \textbf{83.55} & \textbf{87.63} & \underline{88.21} & 94.51 & \textbf{84.26} & \textbf{93.39} & \textbf{96.62} & \textbf{95.99} & \textbf{95.52} & \textbf{93.20} & \textbf{97.71} & \textbf{96.24} & \textbf{90.90} & \textbf{93.49} & \textbf{88.22} & \textbf{91.96}\(_{\pm0.04}\)  \\
    
\bottomrule
\end{tabular}%
}
\label{tab:cifar10c_vits}
\end{table*}

\subsection{Baselines}
\label{app:base-baselines}

To validate the effectiveness of CoDiRe, we compare it against a comprehensive set of baselines, TTA methods, CTTA methods, VLM-TTA methods, and TTD methods. 
Below, we summarize these baselines, along with their configurations in this paper for reproducibility. 

\paragraph{TTA Methods.}
TTA methods include:
\begin{itemize}
    \item \textbf{Source (No Adaptation):} A baseline that directly uses the pre-trained model for inference on test data without any adaptation.
    \item \textbf{BN Adapt:} This method replaces Batch Normalization (BN) statistics with those computed from the current test batch, commonly referred to as Target Batch Normalization (TBN).
    \item \textbf{Tent:} Tent uses entropy minimization as a self-supervised loss to encourage the model to adapt to the target domain.
    \item \textbf{MEMO:} MEMO improves model robustness by averaging probabilities across multiple augmented views of the same input. After adapting to each batch, the model is recovered to the pre-trained, source version, making the adaptation process episodic. The augmentation size is set to 32, and the moving average rate for updating the teacher model is set to 0.999. 
    \item \textbf{EATA:} EATA mitigates noisy gradients by introducing entropy-based filtering and weighting strategies. The entropy threshold $E_0$ is set to $\texttt{log}(K) \times 0.4$, where $K$ is the number of classes, and the threshold $\epsilon$ for filtering redundant samples is set to 0.05. We use fisher regularizer here as default.  
    \item \textbf{SAR:} SAR addresses noisy gradients in challenging scenarios such as small batch sizes and mixed distributions. It uses an entropy threshold $E_0 = \texttt{log}(K) \times 0.4$ and a model recovery threshold $e_m = 0.2$.
    \item \textbf{DeYO:} DeYO combines Pseudo-Label Probability Difference (PLPD) with entropy-based filtering for high-quality sample selection. The entropy threshold $\tau_{\text{Ent}}$, PLPD threshold $\tau_{\text{PLPD}}$, and $Ent_0$ are set to $\texttt{log}(K) \times 0.4$, $\texttt{log}(K) \times 0.5$, and 0.3, respectively. The other configuration is kept the same as the original paper. 
\end{itemize}
\begin{table*}[ht]
\caption{\textbf{Comparison results under corruption scenarios on ImageNet-C with ResNet26 as backbone.} Classification accuracy of the standard ImageNet \(\rightarrow\) ImageNet-C online continual test-time adaptation task while continually adapting to different corruptions at the highest severity 5. The best and second-best results are highlighted in \textbf{bold} and \underline{underlined}, respectively.}
\centering
\resizebox{\textwidth}{!}{%
\begin{tabular}{c|c|ccccccccccccccc|c}
\toprule
\textbf{ImageNet-C} &
\multirow{2}{*}{\textbf{Venue}} &
\multicolumn{3}{c}{\textbf{Noise}} &
\multicolumn{4}{c}{\textbf{Blur}} &
\multicolumn{4}{c}{\textbf{Weather}} &
\multicolumn{4}{c|}{\textbf{Digital}} &
\multirow{2}{*}{\textbf{Avg.}} \\ %\cline{2-16}
\textbf{ResNet26-BN} & & \textit{gauss.} & \textit{shot} & \textit{impul.} & \textit{defoc.} & \textit{glass} & \textit{motion} & \textit{zoom} & \textit{snow} & \textit{frost} & \textit{fog} & \textit{brit.} & \textit{contr.} & \textit{elastic} & \textit{pixel} & \textit{jpeg} &  \\ 
\rowcolor{gray!10} & \multicolumn{15}{c|}{\textit{Time}\hspace{2em}\textemdash\textemdash\textemdash\textemdash\textemdash\textemdash\textemdash\textemdash\textemdash\textemdash\textemdash\textemdash\textemdash\textemdash\textemdash\textemdash\textemdash\textemdash\textemdash\textemdash\textemdash\textemdash\textemdash\textemdash\textemdash\textemdash\textemdash\textemdash\textemdash\textemdash\textemdash\textemdash\textemdash\textemdash\textemdash\textemdash\textemdash\textemdash\textemdash\textemdash\textemdash\textemdash\textemdash>} & \\ 
\midrule
\textbf{Source} 
    & - & 2.06 & 3.04 & 2.58 & 11.66 & 6.86 & 10.74 & 20.72 & 10.56 & 15.38 & 15.94 & 51.78 & 2.60 & 11.20 & 17.76 & 31.36 & 14.28  \\

\textbf{BN Adapt} 
    & NIPS'20 & 11.59 & 13.47 & 11.45 & 10.81 & 12.80 & 20.07 & 32.41 & 27.76 & 28.21 & 38.63 & 58.96 & 7.86 & 38.17 & 38.75 & 26.79 & 25.18\(_{\pm0.07}\)  \\

\textbf{Tent}
    & ICLR'21 & 11.76 & 13.40 & 12.05 & 11.24 & 12.35 & 19.98 & 32.70 & 27.92 & 28.36 & 39.08 & 59.27 & 8.09 & 39.01 & 40.07 & 28.33 & 25.57\(_{\pm0.09}\)  \\

\textbf{MEMO}
    & NIPS'22 & 1.37 & 2.11 & 1.54 & 9.65 & 8.23 & 12.49 & 23.04 & 16.01 & 18.44 & 18.63 & 51.25 & 3.29 & 15.38 & 24.00 & 30.19 & 15.71\(_{\pm0.14}\)  \\

\textbf{EATA}
    & ICML'22 & 10.91 & 13.32 & 11.90 & 10.93 & 12.61 & 19.96 & 33.05 & 28.60 & 29.17 & 40.42 & 60.57 & 9.03 & 39.71 & 41.15 & 29.65 & 26.07\(_{\pm0.09}\)  \\

\textbf{SAR}
    & ICLR'23 & 11.15 & 13.03 & 11.33 & 10.77 & 12.37 & 19.89 & 32.89 & 28.17 & 28.67 & 38.45 & 59.84 & 8.51 & 39.28 & 39.97 & 27.61 & 25.46\(_{\pm0.14}\)  \\

\textbf{DeYO}
    & ICLR'24 & 11.47 & 13.20 & 11.86 & 10.71 & 12.55 & 19.29 & 33.11 & 28.97 & 29.81 & 42.01 & 59.39 & 10.15 & \underline{39.78} & 41.93 & 32.65 & 26.46\(_{\pm0.14}\)  \\

\midrule
\textbf{CoTTA} 
    & CVPR'22 & 11.46 & 12.72 & 11.21 & 9.97 & 11.04 & 16.68 & 30.03 & 26.47 & 26.17 & 35.70 & 59.65 & 7.82 & 37.29 & 38.41 & 25.76 & 24.03\(_{\pm0.08}\)  \\

\textbf{NOTE} 
    & NIPS'22 & 3.88 & 7.66 & 13.60 & 0.09 & 2.53 & 3.16 & 11.99 & 11.62 & 21.53 & 11.22 & 47.89 & 2.48 & 16.79 & 14.87 & 25.82 & 13.01\(_{\pm0.13}\)  \\

\textbf{RoTTA} 
    & CVPR'23 & 9.42 & 14.21 & 13.89 & 9.47 & 11.83 & 16.97 & 30.91 & 13.66 & 23.61 & 30.81 & 56.82 & 3.01 & 25.55 & 15.44 & 25.58 & 20.08\(_{\pm0.36}\)  \\

\textbf{SANTA}
    & TMLR'23 & 11.20 & 13.30 & 11.40 & 11.39 & 12.28 & 19.77 & 32.58 & 28.19 & 29.27 & 38.81 & 59.45 & 8.05 & 38.45 & 39.91 & 28.06 & 25.47\(_{\pm0.12}\)  \\

\textbf{ViDA}
    & ICLR'24 & 11.88 & 13.13 & 11.89 & 10.87 & 12.19 & 19.70 & 33.18 & 27.82 & 29.22 & 40.23 & 59.92 & 9.55 & 39.45 & 41.25 & 30.15 & 26.03\(_{\pm0.04}\)  \\

\midrule
\textbf{CLIP}
  & ICML'25 & \underline{22.94} & \underline{23.20} & \underline{24.06} & \underline{31.50} & \underline{19.80} & \underline{35.84} & \underline{33.58}
  & \underline{45.00} & \underline{39.34} & \underline{47.26} & \underline{62.88} & \underline{34.54} & 25.74 & \underline{50.68} & \underline{42.02} & \underline{35.89} \\

\rowcolor{cyan!10}\textbf{Ours} 
    & - & \textbf{27.36} & \textbf{29.55} & \textbf{30.10} & \textbf{35.59} & \textbf{25.77} & \textbf{42.13} & \textbf{43.97} & \textbf{52.17} & \textbf{48.17} & \textbf{57.73} & \textbf{72.38} & \textbf{39.71} & \textbf{41.95} & \textbf{60.57} & \textbf{52.39} & \textbf{43.97}\(_{\pm0.04}\)  \\
    
\bottomrule
\end{tabular}%
}
\label{tab:imagenet_rn26}
\end{table*}

\paragraph{CTTA Methods.}
CTTA methods include:
\begin{itemize}
    \item \textbf{CoTTA:} A method for handling continuous domain shifts by leveraging temporal consistency and pseudo-label refinement. We set the augmentaion size to 32, the stochastic rate as 0.01, and the weight for the teacher model as 0.999. 
    As we find that CoTTA is hyperparameter-sensitive, we set the confidence threshold to 0.62 in CIFAR-10-C, 0.52 in CIFAR-100-C, 0.1 in ImageNet-C, and 0.5 in other datasets.  
    \item \textbf{NOTE:} Addresses temporally correlated test streams via Instance-Aware Batch Normalization (IABN) and Prediction-Balanced Reservoir Sampling (PBRS), which simulates i.i.d. sampling from non-i.i.d. streams using a memory bank.
    \item \textbf{RoTTA:} A method that tackles dynamic distribution shifts via robust statistics estimation and Category-balanced Sampling with Timeliness and Uncertainty (CSTU). We use the Adam optimizer here, and replace the standard Robust Batch Normalization (RBN) layers with learnable LN layers for ViT backbones. 
    \item \textbf{SANTA:} A source anchoring network designed for target alignment in dynamic environments. As the original version of SANTA requires access to the source domain samples, we use a dynamically updated version of the prototype instead. 
    \item \textbf{ViDA:} This approach focuses on homeostatic adaptation to balance stability and plasticity during continuous adaptation. All the configuration is kept the same as the original paper. 
    \item \textbf{DPCore:} DPCore manages domain knowledge via a dynamic prompt coreset, utilizing Visual Prompt Adaptation (VPA) to align domains efficiently. We use the AdamW optimizer, with the prompt length set to 8 and the update threshold $\rho$ set to 0.8.
\end{itemize}

\paragraph{VLM-TTA Methods.}
VLM-TTA methods include:
\begin{itemize}
    \item \textbf{TPT:} A method that uses prompt tuning to improve generalization during test-time adaptation. We set the augmentation size to 64, and confidence threshold $\rho$ to 0.1. 
    \item \textbf{TDA:} A training-free approach that dynamically updates a cache with high-quality features from historical test data. All the configuration is kept the same as the original paper.
    \item \textbf{BoostAdapter:} An adapter-based approach that bootstraps VLM performance through regional feature refinement. All the configuration is kept the same as the original paper.
    \item \textbf{ZERO:} A training-free approach that simply zeroing out the Softmax temperature over augmented views and performing majority voting on confident predictions. All the configuration is kept the same as the original paper.
\end{itemize}

\paragraph{TTD Methods.}
TTD methods refer to the variants under our proposed Test-Time Distillation paradigm. They include:
\begin{itemize}
    \item \textbf{Naive Ensemble:} A baseline that computes the inference output by simply averaging the logits of the target model and the frozen CLIP without updating any parameters.
    \item \textbf{BN Adapt w. NE:} This method performs BN Adapt (updating BN statistics on the test stream) on the target model, and uses the Naive Ensemble strategy for the final inference output.
    \item \textbf{Tent w. NE:} This method performs Tent (entropy minimization) on the target model parameters, and utilizes the Naive Ensemble strategy for the final inference output.
    \item \textbf{Distill CLIP:} A direct distillation baseline where the target model is updated to match the predictions of the frozen CLIP model. 
\end{itemize}

\subsection{Hyperparameters}
\label{app:base-hyper}

To ensure fair and reproducible comparisons, we use consistent experimental settings across all methods. Below, we summarize the hyperparameters for each dataset and scenario:

\begin{itemize}
    \item \textbf{CIFAR-10-C and CIFAR-100-C:} We use ResNet50 as the default backbone of target model, and also provide the results of ResNet26 and ViT/S-16 in Appendix \ref{app:exp}. The learning rate is set to $1 \times 10^{-4}$. 
    \item \textbf{ImageNet-C:} ViT-B/16 is used as the backbone of target model. Similarly, the results of ResNet26 and ResNet50 are also supplemented in Appendix \ref{app:exp}. The learning rate is set to $1 \times 10^{-3}$ in ResNet architectures and $2.5\times 10^{-4}$ in ViT architectures. 
    \item \textbf{PACS and Office-Home:} ResNet50 is used as the backbone of target model, with a learning rate of $1 \times 10^{-4}$.
\end{itemize}

\section{Implementation Details}
All experiments are conduced on the TTAB framework for fair evaluation, run on a single NVIDIA Tesla V100 GPU, and repeated three times with random seeds in the ranges [2022, 2023, 2024]. 
All the baselines' algorithmic settings (e.g., hyper-parameters) are set to their default values unless otherwise specified. 
As for our method, we set the reset threshold $\gamma_0$ to 0.25, step size $s$ to 20 steps, and reset ratio $\alpha$ to 20(\%).
Unless otherwise specified, all experiments were conducted using SGD as the optimizer, with the batch size set to 64.
Source code used in this paper is under the MIT License. 

\section{Additional Experiments and Analyses}
\label{app:exp}
To further validate the robustness and versatility of CoDiRe, this section provides additional comparative results across different datasets and variations in the target model's architecture. 
These experiments supplement the main paper's findings, offering a broader view of CoDiRe's performance against baseline methods under diverse conditions.

\begin{table*}[ht]
\caption{\textbf{Comparison results under corruption scenarios on ImageNet-C with ResNet50 as backbone.} Classification accuracy of the standard ImageNet \(\rightarrow\) ImageNet-C online continual test-time adaptation task while continually adapting to different corruptions at the highest severity 5. The best and second-best results are highlighted in \textbf{bold} and \underline{underlined}, respectively.}
\centering
\resizebox{\textwidth}{!}{%
\begin{tabular}{c|c|ccccccccccccccc|c}
\toprule
\textbf{ImageNet-C} &
\multirow{2}{*}{\textbf{Venue}} &
\multicolumn{3}{c}{\textbf{Noise}} &
\multicolumn{4}{c}{\textbf{Blur}} &
\multicolumn{4}{c}{\textbf{Weather}} &
\multicolumn{4}{c|}{\textbf{Digital}} &
\multirow{2}{*}{\textbf{Avg.}} \\ %\cline{2-16}
\textbf{ResNet50-BN} & & \textit{gauss.} & \textit{shot} & \textit{impul.} & \textit{defoc.} & \textit{glass} & \textit{motion} & \textit{zoom} & \textit{snow} & \textit{frost} & \textit{fog} & \textit{brit.} & \textit{contr.} & \textit{elastic} & \textit{pixel} & \textit{jpeg} &  \\ 
\rowcolor{gray!10} & \multicolumn{15}{c|}{\textit{Time}\hspace{2em}\textemdash\textemdash\textemdash\textemdash\textemdash\textemdash\textemdash\textemdash\textemdash\textemdash\textemdash\textemdash\textemdash\textemdash\textemdash\textemdash\textemdash\textemdash\textemdash\textemdash\textemdash\textemdash\textemdash\textemdash\textemdash\textemdash\textemdash\textemdash\textemdash\textemdash\textemdash\textemdash\textemdash\textemdash\textemdash\textemdash\textemdash\textemdash\textemdash\textemdash\textemdash\textemdash\textemdash>} & \\ 
\midrule
\textbf{Source} 
    & - & 18.32 & 18.88 & 17.70 & 16.72 & 8.70 & 19.02 & 25.76 & 27.64 & 30.00 & 35.22 & 63.36 & 20.94 & 12.16 & 25.58 & \underline{47.88} & 25.86 \\

\textbf{BN Adapt} 
    & NIPS'20 & 10.12 & 12.36 & 10.18 & 9.70 & 10.10 & 20.28 & 34.46 & 35.50 & 38.06 & 53.18 & 65.88 & 16.28 & 36.66 & 34.82 & 30.74 & 27.89\(_{\pm0.03}\)  \\

\textbf{Tent}
    & ICLR'21 & 10.73 & 13.75 & 11.26 & 9.92 & 11.18 & 21.09 & 34.66 & 36.71 & 40.97 & \underline{55.73} & \underline{66.64} & 19.71 & 39.39 & 37.51 & 37.86 & 29.81\(_{\pm0.20}\)  \\

\textbf{MEMO}
    & NIPS'22 & 14.47 & 16.67 & 14.83 & 13.24 & 11.14 & 18.79 & 26.99 & 31.86 & 32.87 & 23.12 & 61.01 & 2.39 & 19.71 & 31.43 & 46.34 & 24.32\(_{\pm0.02}\)  \\

\textbf{EATA}
    & ICML'22 & 10.32 & 14.11 & 12.37 & 10.26 & 11.68 & 20.55 & 33.90 & 37.31 & 40.08 & 52.98 & 64.45 & 22.59 & 40.08 & 41.89 & 41.55 & 30.28\(_{\pm1.06}\)  \\

\textbf{SAR}
    & ICLR'23 & 11.15 & 13.83 & 11.09 & 9.79 & 11.03 & 19.90 & 34.83 & 37.44 & 39.73 & 53.85 & 66.01 & 19.29 & 38.79 & 38.21 & 36.01 & 29.40\(_{\pm0.01}\)  \\

\textbf{DeYO}
    & ICLR'24 & 9.44 & 14.06 & 12.40 & 10.52 & 12.42 & 21.28 & \underline{35.22} & 38.02 & \underline{42.10} & 54.40 & 64.54 & 24.72 & \underline{41.92} & 42.34 & 41.76 & 31.01\(_{\pm0.03}\)  \\

\midrule
\textbf{CoTTA} 
    & CVPR'22 & 11.04 & 13.68 & 11.41 & 9.81 & 10.27 & 20.30 & 33.82 & 35.87 & 38.25 & 53.87 & 65.21 & 16.22 & 37.24 & 34.81 & 31.63 & 28.23\(_{\pm0.12}\)  \\

\textbf{NOTE} 
    & NIPS'22 & 22.76 & 21.65 & 22.33 & 0.16 & 2.61 & 6.20 & 18.98 & 23.91 & 39.02 & 26.41 & 62.79 & 17.09 & 17.41 & 13.49 & 33.33 & 21.88\(_{\pm0.24}\)  \\

\textbf{RoTTA} 
    & CVPR'23 & 12.13 & 11.94 & 11.91 & 1.59 & 4.89 & 11.87 & 32.90 & 30.25 & 38.86 & 45.83 & 64.10 & 18.98 & 25.49 & 26.25 & 33.68 & 24.71\(_{\pm0.73}\)  \\

\textbf{SANTA}
    & TMLR'23 & 10.42 & 14.21 & 10.65 & 9.64 & 10.97 & 20.87 & 35.18 & 37.12 & 40.59 & 55.15 & 65.75 & 18.38 & 39.03 & 36.79 & 36.01 & 29.38\(_{\pm0.08}\)  \\

\textbf{ViDA}
    & ICLR'24 & 10.65 & 13.67 & 10.45 & 10.08 & 10.77 & 20.97 & 34.21 & 36.42 & 38.95 & 54.42 & 65.89 & 17.83 & 37.65 & 34.49 & 32.71 & 28.61\(_{\pm0.12}\)  \\

\midrule
\textbf{CLIP}
  & ICML'21 & \underline{22.94} & \underline{23.20} & \underline{24.06} & \underline{31.50} & \underline{19.80} & \underline{35.84} & 33.58
  & \underline{45.00} & 39.34 & 47.26 & 62.88 & \underline{34.54} & 25.74 & \underline{50.68} & 42.02 & \underline{35.89} \\

\rowcolor{cyan!10}\textbf{Ours} 
    & - & \textbf{27.57} & \textbf{31.39} & \textbf{32.01} & \textbf{33.74} & \textbf{27.13} & \textbf{42.20} & \textbf{48.11} & \textbf{56.02} & \textbf{55.20} & \textbf{64.28} & \textbf{75.19} & \textbf{44.03} & \textbf{50.61} & \textbf{62.15} & \textbf{57.81} & \textbf{47.16}\(_{\pm0.47}\)  \\

\bottomrule
\end{tabular}%
}
\label{tab:imagenet_rn50}
\end{table*}

\subsection{More Comparison Results}
\paragraph{Comparison Results on CIFAR-100-C.}
\label{app:cifar100c}
We extend our evaluation to the CIFAR-100-C benchmark, which presents a more challenging corruption adaptation task due to its significantly larger number of classes compared to CIFAR-10-C. Table~\ref{tab:cifar100c} details the performance of CoDiRe against various TTA and CTTA methods, using ResNet50 as the backbone. 
The results unequivocally demonstrate CoDiRe's superior robustness, maintaining high accuracy even in this complex scenario where baselines struggle to distinguish between fine-grained categories.

\paragraph{More Comparison Results on CIFAR-10-C.}
To verify the model-agnostic nature of CoDiRe, we assess its performance on CIFAR-10-C using different backbones for the target model. 
Table~\ref{tab:cifar10c_rn26} shows the substantial gains with a ResNet26 backbone, while Table~\ref{tab:cifar10c_vits} presents results with a ViT-S/16 backbone. 
These outcomes confirm that CoDiRe is not reliant on specific architectures and consistently delivers impressive improvements across varying model capacities.

\paragraph{More Comparison Results on ImageNet-C.}
Similarly, we demonstrate the seamless scalability of our method on the large-scale ImageNet-C benchmark. Table~\ref{tab:imagenet_rn26} details performance using ResNet26, and Table~\ref{tab:imagenet_rn50} provides results when employing ResNet50, complementing the ViT-B/16 results presented in the main paper. 
These results confirm CoDiRe's consistent superiority and stability, proving its effectiveness regardless of the backbone architecture size on challenging large-scale benchmarks. 
Moreover, the smaller the target model is, the greater it can benefit from distillation. This is quite natural phenomenon.

\end{document}